\documentclass[sigconf, nonacm,authorversion]{acmart}

\usepackage[ruled,linesnumbered]{algorithm2e}

\usepackage{graphicx}
\usepackage{textcomp}
\usepackage{amsmath}
\usepackage{array}
\usepackage{stfloats}
\usepackage[bottom]{footmisc}
\usepackage{subcaption}

\usepackage{xcolor}

\usepackage[nolist,nohyperlinks]{acronym}
\acrodef{CR}[$C_R$]{\emph{Contrast Ratio}}
\acrodef{CSF}[CSF]{\emph{Contrast Sensitivity Function}}
\acrodef{LoS}[LoS]{\emph{Line of Sight}}

\newcommand{\eg}{\textit{e.g.},\xspace}
\newcommand{\ie}{\textit{i.e.},\xspace}

\AtBeginDocument{%
  \providecommand\BibTeX{{%
    \normalfont B\kern-0.5em{\scshape i\kern-0.25em b}\kern-0.8em\TeX}}}

\acmConference[SimAUD '21]{SimAUD '21: Symposium on Simulation in Architecture and Urban Design}{April 15--17, 2021}{}

\begin{document}

\title{Adding Visibility to Visibility Graphs: Weighting Visibility Analysis with Attenuation Coefficients}

\author{Mathew Schwartz}
\email{cadop@njit.edu}
\orcid{0000-0003-3662-7203}
\affiliation{%
  \institution{New Jersey Institute of Technology}
  \city{Newark}
  \state{NJ}
  \country{USA}
}
\author{Margarita Vinnikov}
\email{vinnikov@njit.edu}
\affiliation{%
  \institution{New Jersey Institute of Technology}
  \city{Newark}
  \state{NJ}
  \country{USA}
}
\author{John Federici}
\email{john.f.federici@njit.edu}
\affiliation{%
  \institution{New Jersey Institute of Technology}
  \city{Newark}
  \state{NJ}
  \country{USA}
}

\renewcommand{\shortauthors}{Schwartz, et al.}

\begin{abstract}
Evaluating the built environment based on visibility has been long used as a tool for human-centric design. The origins of isovists and visibility graphs are within interior spaces, while more recently, these evaluation techniques have been applied in the urban context. One of the key differentiators of an outside environment is the weather, which has largely been ignored in the design computation and space-syntax research areas. While a visibility graph is a straightforward metric for determining connectivity between regions of space through a line of sight calculation, this approach largely ignores the actual visibility of one point to another. This paper introduces a new method for weighting a visibility graph based on weather conditions (\ie rain, fog, snow). These new factors are integrated into visibility graphs and applied to sample environments to demonstrate the variance between assuming a straight line of sight and reduced visibility. 
\end{abstract}



\begin{teaserfigure}
  \includegraphics[page=11, width=\textwidth]{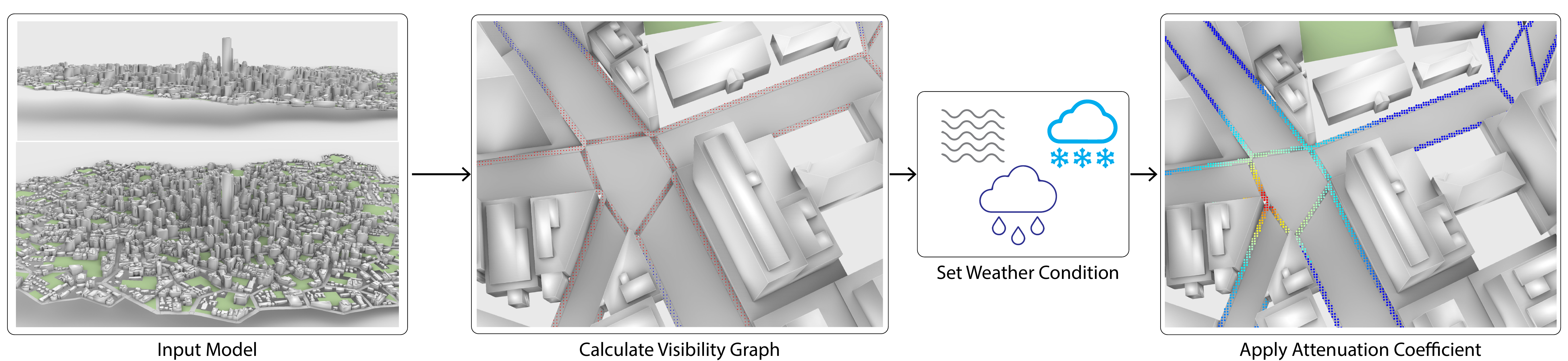}
  \caption{Overview of framework. An input model is provided, which we calculate the visibility graph for. The user then sets the weather condition(s). We apply attenuation coefficients calculated per weather condition on the edge distances and re-weight the visibility graph.}
  \Description{Visibility graph using attenuation coefficients from weather.}
  \label{fig:teaser}
\end{teaserfigure}

\maketitle

\section{Introduction}

\begin{figure}[!htbp]
\centering
\includegraphics[page=4, width=.45\textwidth]{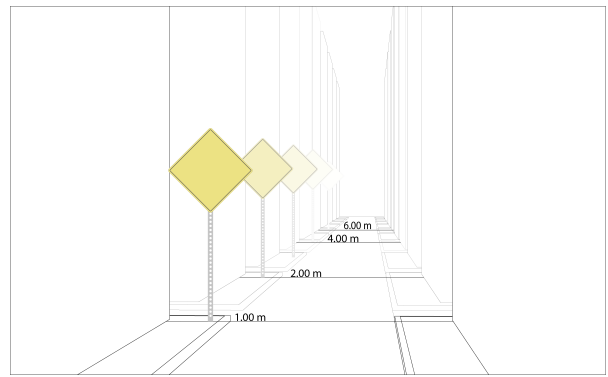}
\caption{Illustration of the fall off from a view in an urban environment in which the weather conditions makes the distant objects non-visible.  }
\label{fig:visibilityConcept} 
\end{figure}

When translating the experience a person has of large open halls and high ceilings of a building to the urban scale, there may be an assumption that these spatial experiences translate well, as humans remain the same. However, at large distances, atmosphere and environmental conditions impact visibility (Fig.~\ref{fig:visibilityConcept}). 
The intuition and results gained from spatial analysis and human subject studies at the building scale (\eg~\cite{choi1999morphology}) and, in many cases of urban environments, miss the impact of adverse weather. Depending on the region, the extent to which these adverse weather conditions happen may vary. However, times in which adverse weather conditions exist may very well be when visibility matters the most.  Furthermore, the increased rate of extreme weather patterns from climate change~\cite{fischer2016observed} suggests, this consideration will not be unnecessary any time soon.  

The importance of considering how adverse weather impacts visibility--and therefore, human experience in the built environment--is tied not only to the visceral experience, but to proper wayfinding and navigational cues as well (\eg Fig.~\ref{fig:visibilityConcept}). While adverse weather is not the status quo in many regions, it is important to consider the human experience of the built environment not only at the optimal condition of what we (as designers) \textit{intend} for the possibility of a design, but to the very real problems and conditions of the environment as well. The importance of this work is considered in light of the frequency adverse weather occurs, just as one would hope a building uses fire-retardant materials and properly lit egress signs, \ie; not for the expectation of commonly occurring fires, but simply to the possibility. From a broad perspective, people's ability to efficiently navigate an environment relies on visible cues, such as the signs indicating a bus stop location. As research shows reduced use of public transportation during adverse weather~\cite{miao2019extreme} (albeit not specific to visibility), improving the users' ability to easily find transportation stops with coverings is one more step in improving accessibility. 

In this paper, we introduce the methods of calculating visibility based on a variety of weather conditions, use these calculations for scoring edges of a graph and view-based distance, and demonstrate how these new metrics perform in spatial analysis along with the difference in a simple space-syntax metric. Unlike a commonly used calculation of visibility referring to a set cut-off threshold, the method implemented in this paper uses a range from which the cut-off threshold is derived. Section~\ref{subsec:sig} introduces the rationale and significance of visibility thresholds and contrast, as well as the role visibility has in understanding design options. Section~\ref{sec:methods} details the calculations used for weighting edges. Finally, section~\ref{sec:results} demonstrates the weighted visibility measures on traditional case studies to explain visibility graphs and large-scale virtual cities.  

\section{Significance and Related Work}\label{subsec:sig}

The following sections offer a brief background into the concept of visibility and visibility analysis from the physiological and design perspectives to provide context to our work. First, we present a rationale for why we diverge from the traditional implementation of visibility as a threshold value. Second, we present our work within the context of a design discipline. 

\subsection{Visibility}\label{subsec:vis}

\begin{figure}[!htbp]
\centering
\includegraphics[width=.45\textwidth]{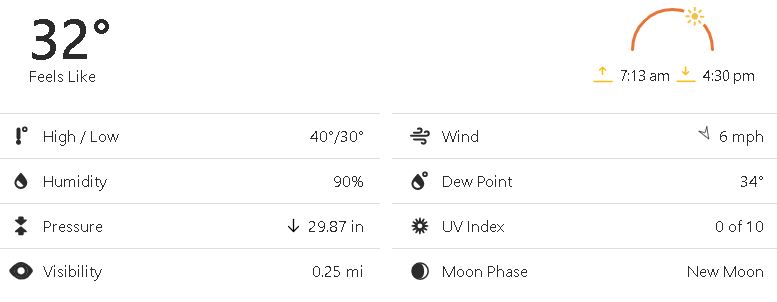}
\caption{ A screenshot of \url{weather.com}. Bottom left of the image shows \textit{Visibility} as 0.25 mi.}
\label{fig:weather.com} 
\end{figure}

Visibility is a function of the distance and attenuation coefficient when light passes through a uniform media such as the atmosphere. The term is used when reporting the maximum distance one can see the difference in maximum contrast in a particular light and weather condition. For example, this relationship between distance and attenuation is used in reference to weather reports (\eg 0.25mi  \url{weather.com}, Fig.~\ref{fig:weather.com}),
where it is used to show the distance at which an object or light can be clearly identified. The value shown in Figure~\ref{fig:weather.com} derives from the Koschmieder equation(Eq.~\ref{eq:koshmieder}):

\begin{equation}\label{eq:koshmieder}
    V_d = \frac{3.912}{\sigma}
\end{equation}
where $V_d$ is the visibility distance and $3.912$ is derived from the $2\%$ contrast ratio as: $\texttt{ln}(0.02) = -3.912$. However, the use of the Koschmieder equation as a reference to visibility can be confusing in the context of the built environment and design, where the \textit{visibility} of something is unlikely to be considered in absolutes. Rather, this reference of visibility with a specific distance should be considered as object detection and not necessarily identification~\cite{lee2016visibility}. For example, an occupant may be able to detect an object is in the distance, but not that it is a sign with information on it (Fig.~\ref{fig:visibilityConcept}). The original equation, the logarithm of a percentage, defines the \ac{CR}. The visual contrast within an environment is determined by the light difference between an observer from the background and a black object. This value, reported as a ratio (referred to here as \ac{CR}), is shown in Equation~\ref{eq:VisSigmaDistance}:

\begin{equation}
 C_r = \texttt{exp}(-\sigma x) \label{eq:VisSigmaDistance}
\end{equation}
where $\sigma$ is an attenuation coefficient, and $x$ is a distance. As seen, the Koschmieder equation assumes a contrast ratio of $2\%$. With this assumption, the distance an object is visible ($x=V_d$) is given by only supplying $\sigma$. 

Original visibility calculations of Kruse et al. \cite{kruse1962elements} used $2\%$ as a threshold, while later $5\%$ was suggested as a better estimate and used for runway visibility~\cite{gordon1979daytime}. While the validity of these values are still being addressed in recent literature~\cite{lee2016visibility}, these types of changes reinforce the value of using visibility metrics that are computationally varied by distance rather than a predetermined value. Nonetheless, in real-life situations, $2\%$ or even $5\%$ are very small values, as discussed in Section~\ref{subsec:physio}, as humans can perceive very few objects at such low contrast. It is important to understand how and at what distance objects, signs, and targets are visible at any ratio, enabling context-specific decisions to be made on what values are relevant. 

Complicating the matter further, as this is a matter of contrast, a dense forest or urban environment is unlikely to have any pure white background and black objects contrasting with each other, making the cutoff of visibility disjointed from the reality of an environment. The driving principle for analysis methods in the built environment is the experience of the occupant. For this reason, two important criteria considered in this paper are 1) locations for analysis are accessible locations of an occupant and 2) the analysis method is linked to a physiological response (human-centric). 

\subsection{Physiology of Visibility}\label{subsec:physio}
While the human visual system is robust and sophisticated, it has some limits in capabilities. There is a limit on the amount of light intensity that an observer can perceive. Specifically, there is a threshold at which observers can no longer identify an object from its background or to resolve detail within an observed object. This is referred to as \textit{visual threshold}. To find such a threshold, psychophysical researchers present participants with stimuli at various levels of contrasts, measuring at what level participants can detect the stimuli 50 percent of the time the stimuli were shown. There are three types of thresholds - spatial, temporal, and color. The spatial threshold can be measured as visual acuity or threshold luminance (the intensity of light reflected off objects) contrast. Temporal contrast sensitivity is measured as sensitivity to contrast as function of time. The assumption for spatial and temporal thresholds that the visual stimuli is grey, while in reality people see a variation of colors. Furthermore, color thresholds are individualized and are not fully standardized. Consequently, will be the focus of our work.

\textit{Visual acuity} is a measure of the smallest detail a person can resolve (provided that luminance contrast is fixed and optimal). In other words, what level of details one sees when presented with black on white symbols.  Visual acuity is expressed in minutes of arc, where $1 arcmin = \frac{1}{60}$ of a degree. One of the most common examples of a visual acuity test is a \emph{Snellen} test, commonly used for eye exams, where letters of different sizes are displayed and the subject needs to correctly identify the letters. Hence the acuity is defined as follows:

\begin{equation}
Acuity = \frac{1}{gap size} arc min
\end{equation}

This is, in particular, important to understand in the context of signage. The usual assumption is that people with normal or corrected to normal vision can resolve targets at 1 min arc. Where \emph{corrected to normal vision} or \emph{20/20 vision} term is used to describe a visual capability of an average person. Meaning that people at 20 feet (6 m) can see content that is "normally" (by others) seen at this distance. Any other variation implies how people deviate from an average observer. For example, 20/10 vision would mean that people at 20 feet would see what a person with normal vision would see at 10 feet. Hence, it is an important measure to understand what an average person can see at what distance, yet, it also important to understand that some people, especially the elderly--might not be able to resolve targets that an average person can see. 

Furthermore, in real-world scenarios, people do not see black and white. They also see various gradations of grey. Hence, our analysis of the human perception should include both high and mid low-contrast conditions. Therefore, \emph{Contrast sensitivity}, on the other hand, depends on both the \emph{contrast} and the \emph{spatial frequency} (relative size of a stimulus). There are different ways to measure contrast. For example, \emph{Weber's Law} predicts the minimum detectable difference in luminance between test spots on a uniform visual field. Weber contrast is defined as:

\begin{equation}
   \frac{I-I_\mathrm{b}}{I_\mathrm{b}}
\end{equation}
where $I$ is the luminance of the features and $I_{ \mathrm{b} }$ is the luminance of the background. Yet, it is usually used in cases where small features are present on a large uniform background, since outside of these areas, the threshold sensitivity drops off significantly~\cite{luebke2003level}. 

To summarize; \textit{Visual acuity} measures visibility at the high contrast range, \textit{visibility graphs} measure visibility at the highest contrast range, yet \textit{visibility} measures only the lowest contrast range. Furthermore, visibility analysis is not standardized and needs to consider various factors such as luminance between targets and background, targets' size, viewing location and duration, as well as physiological capabilities of an observer such as viewer age. Therefore, we introduce a method of using the \textit{contrast ratio in visibility graphs} that can handle a wide range of metrics. This method can be used then by architects, city planner, and general population, when they need to consider alternatives to the contrast threshold for determining visibility under particular conditions. In particular, our method is addressing the issue of \textit{visual range} that defines the range of distances at which a target can be seen during weather and the atmospheric effects of light extinction.

\subsection{Analysis of the Built Environment}\label{subsec:env}

\begin{figure*}[!htbp]
\centering
    \begin{subfigure}[b]{0.49\textwidth}
        \includegraphics[page=1, width=\textwidth]{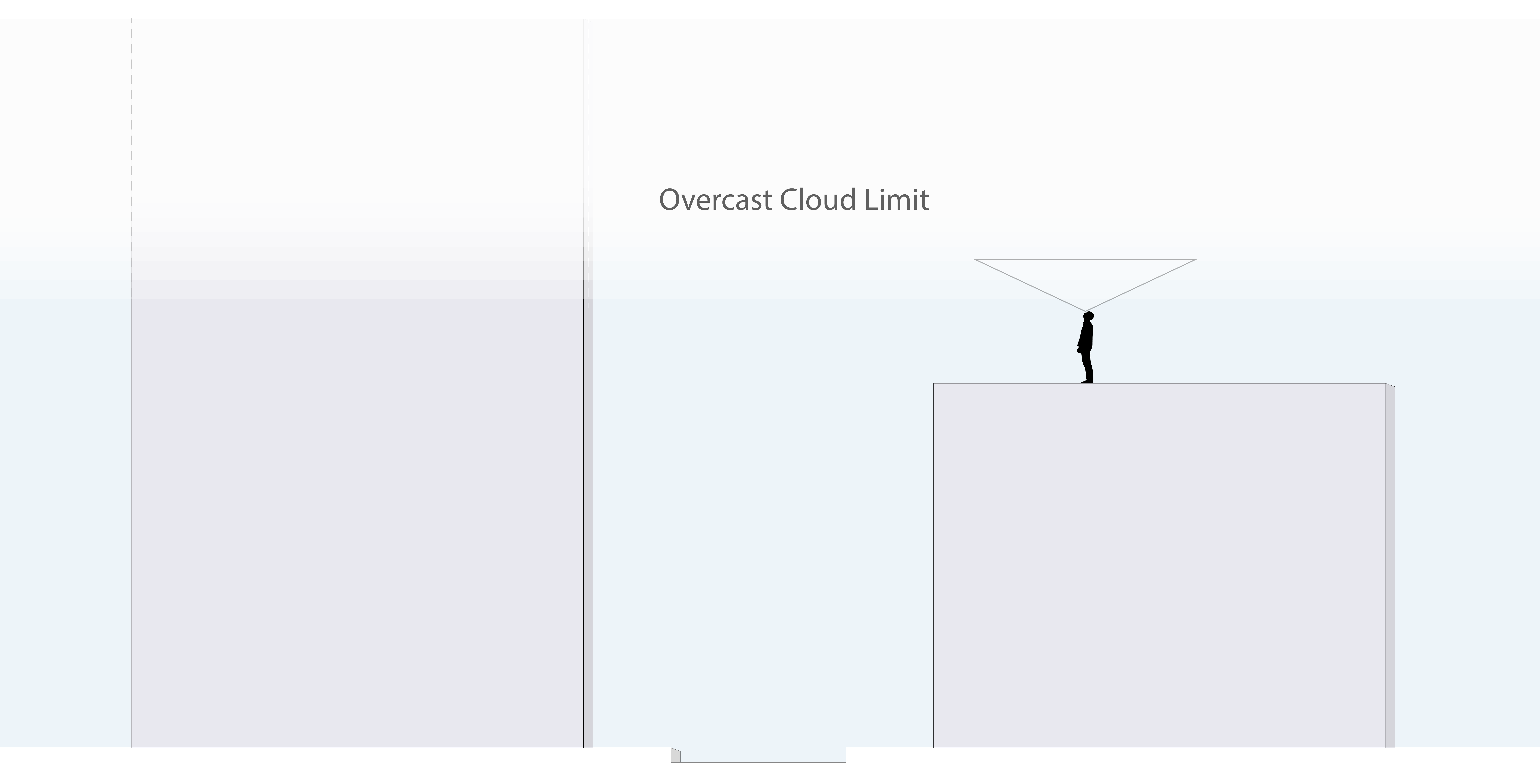}
        \caption{Looking into the sky on overcast day. }
        \label{fig:skyview}
    \end{subfigure}
    \begin{subfigure}[b]{0.49\textwidth}
        \includegraphics[page=2, width=\textwidth]{figures/CR_Concepts.pdf}
        \caption{Adverse weather visual range of buildings on sidewalk.}
        \label{fig:buildingview}
    \end{subfigure}
\caption{When considering an overcast sky, the contrast reduction of the sky by clouds creates a blank view for an occupant (\ref{fig:skyview}), standing on the top of the building on the right, looking up (field of view shown as a triangle). This view is near identical to that of a person looking across the street to a building with adverse weather conditions. At the same time, they would be able to see the building and/or information associated (\eg signage) when close (\ref{fig:buildingview}), represented by two people on a sidewalk facing the left and right buildings. }
\label{fig:weatherConcept} 
\end{figure*}

Spatial analysis is often performed on a graph that ranges in the way it is constructed and what it represents. A network of lines that create direct connections between key points of interest is often used for route and path planning for both indoor (e.g.,~\cite{lee2010computing,shin2019indoor,suter2013structure}) and outdoor (e.g.,~\cite{peroche2014accessibility,Fuchkina2017}) environments. When considering the analysis of an entire space, and not just of the paths, all possible locations within a model must be categorized as accessible or not. There are a range of methods in the literature for determining which locations of an environment are human-accessible, ranging from surface-defined regions~\cite{lamarche2009topoplan,pettre2005navigation} to grid-based graph of points~\cite{nagy2017buzz,Schwartz:2019:nonflat} or both \cite{kallmann2014navigation}. In addition to the human movement aspect of the dense grid, items or objects in the environment can be associated directly to the closest node. For example, the entrance of a building can be analyzed by the nearest node in the graph, or specific nodes of \textit{good} visibility can be associated with walls or street corners for sign placement. 

The representations of visibility within the built environment that our work is built upon are isovists, visibility graphs, and viewsheds. Early work in visibility analysis of space was in the use of isovists~\cite{benedikt1979take}, which are polygons representing a field of view from a given location (a node). The intersecting locations of polygons can then be reduced to numeric values based on various metrics. Around the same time as the introduction of isovists in spatial analysis, the concept of a visibility graph was also introduced~\cite{lozano1979algorithm}, in which a graph is constructed from the direct connections of all vertices in a set given geometric obstacles. This allows for optimal shortest path planning within an environment. The direct connections of a visibility graph for path planning were re-purposed for the built environment and spatial analysis in~\cite{turner2001isovists} by connecting a generated grid of nodes along a floor surface. While these two analysis methods are distinct in origin, there is considerable overlap in the use and metrics calculated. 

Both isovists and visibility graphs have been continuously developed in the literature for a variety of implementation methods and new metrics to be incorporated. ~\cite{conroy2001spatial} provides descriptions on numerous metrics that can be calculated with isovist arrays, and in turn, a visibility graph. While distance is referred to and used, a majority of past research has focused on the graph distance (number of connections) rather than metric distance. In~\cite{varoudis2014beyond}, this concept is extended to 3D with a \textit{3D visual integration} metric. A 3D visibility graph was also proposed by~\cite{lu2019three}, with the addition of a target point of interest. In traditional isovist literature, often the only reference to distance is in reference to the minimum, maximum, and average extents of the isovist of a node (also referred to as the radial length). Within more generalized visibility analysis there have been examples of using distance as part of the spatial analysis (e.g., ~\cite{koltsova2013visibility,lu2019three,fisher2018integrating,nutsford2015personalising}). 

The need for additional metrics to more accurately represent the human experience is seen in recent literature. The influence of direct connections between nodes that are not accessible directly (\eg seeing a node through a window) is shown in~\cite{varoudis2015visibility}. In \textit{viewshed analysis} implemented with \ac{LoS} methods, ~\cite{nutsford2015personalising} argued for a more human-centric metric incorporating distance and angle within a 3D environment. A 3D \ac{LoS} presented in~\cite{fisher2018integrating,fisher20143d} records the distance and object type for a depth map out of a window view. Occupants view has also been used in evaluating urban space for path planning \textit{w.r.t.} shadows and sun glare during jogging~\cite{Schwartz:jogging}, and as a comparative analysis to perceived density \textit{w.r.t.} the \ac{LoS} to buildings, sky, and nature while considering pedestrian movement~\cite{fisher2017can}. Path analysis considering human energy expenditure and effort on uneven terrain was shown in~\cite{Schwartz:2019:nonflat}, and considered for cognitive models as a function of visible path slope in~\cite{greenberg2020physical}. 

As shown in past research, there is a correlation between the integration of space and the number of people present~\cite{choi1999morphology}. The inverse of this would be that a condition in which spaces that are further away are not visible would reduce integration. In this scenario, the openness of an urban environment or large distances between buildings would be inversely impacted by the traditional isovist metrics. The impact of visibility through space, removed from the built environment, poses a unique challenge for the common metrics and their interpretations. To describe the conceptual problem with visibility analysis as a function of environmental condition, we introduce Fig.~\ref{fig:weatherConcept}. Suppose a person is located one meter from and facing towards a wall. In this instance, the minimum radial length of an isovist would be recorded to reflect this up-close object. If this person is on a roof looking up on an overcast day, their view would contain nothing, while having an infinite minimum radial length. 

\section{Methodology}\label{sec:methods}

To evaluate a location in space for visibility, a \ac{LoS} calculation is done through ray-casting in a virtual 3D environment,  which returns the distance to the closest intersection point, with the ray considered a weighted edge of a graph or as a node attribute. These weights are determined per-condition, such that the distance is used to compute a particular score given an environmental condition. The calculation for each condition is described in Section~\ref{sec:atten_coefs}.

\subsection{Visibility Graph}\label{sec:visgraph}

The visibility graph is defined by the all-to-all connection check of vertices\footnote{although at times this is confusing, vertices of the graph refer to a graph vertex, not a vertex of a triangle or mesh} in the environment. This visibility graph is undirected weighted graph defined as $VG = \{V,E,W\}$, where $V$ is a set of vertices; $v | v_i \in V$, $E$ is the set of edges $e_{ij} \in E$, and $W$ is the set of weights mapping to an edge $w(e_{ij}) \in W ~|~ w(e_{ij}) \rightarrow e_{ij}$. An edge $e_{ij} \in E$ is an unordered pair of vertices $(v_i,v_j)$ (where $v_i \ne v_j$ and $e_{ij} = e_{ji}$) that are defined as having a direct connection with no obstructions in the environment. As the focus of our work is in the visibility (specifically \ac{CR}), a function of distance, $W$ is an important set for analysis. 

Additionally, a directed, or subset $VG$ can be calculated---such that a single vertex, or group of vertices, is considered. In this case, two sets are defined: ${}^{a}V$ is the set of vertices to be considered in the graph, and ${}^{b}V$ is a set of vertices that are used for the \ac{LoS} calculations. Similarly ${}^{a}V \cup {}^{b}V \subset V(VG)$. This relationship can also be understood by the definition of edges:

\begin{equation}
    E(VG) = \{ e_{ij} | v_i \in {}^{a}V \land v_j \in {}^{b}V \}
\end{equation}

There are multiple methods to analyze a $VG$. First, one can consider the attributes on the structure of the graph. The degree $deg(v_i)$ is the number of edges in the graph containing that vertex, representing the number of visible locations from and to that vertex, referred to as \textit{connectivity}. This creates a score for each vertex in the graph that represents \textit{perfect visibility} between connected vertices, as the binary value is $1$ (connection exists) or $0$ (no connection exists). In addition to the degree, a neighborhood of a vertex is defined as the vertices that are connected by an edge, denoted $N_{VG}(v_i) = \{ v_j \in V  | (v_i,v_j) \in E \}$. 

For each condition, the weight mapping is a function of this weather condition, \eg, $w_{ij}(s)$ is the weight mapping for snow, $w_{f_L}(e_{ij})$ light fog, $w_{f_H}(e_{ij})$ heavy fog, and $w_{r}(e_{ij})$ rain. Therefore, the weight of an edge $e_{ij}$ is:

\begin{equation}
   w(e_{ij}) = \texttt{exp}(-\sigma (|\vec{v_i} - \vec{v_j}|))
\end{equation}
where $\vec{v_i}$ and $\vec{v_j}$ are the 3D locations in the environment of the graph vertices. 

Essentially, there must be a reduction from the edge connections that contain a specific vertex, to a single score (that considers these edges) to assigned as an attribute of the vertex itself. $deg(v_i)$ is analogous to the edge count, while also summations $S_S$ and averages $S_A$ can be used on the weight. For summation, the equation is as follows:

\begin{equation}
   S_S(v) = \sum_{x \in N_{VG}(v)} w(e_{vx})
\end{equation}
and the average is:

\begin{equation}
   S_A(v) = \frac{1}{|N_{VG}(v)|}\sum_{x \in N_{VG}(v)} w(e_{vx})
\end{equation}

\subsection{Attenuation Coefficients}\label{sec:atten_coefs}

In the case of spatial analysis, the predetermined distance and contrast ratio of 2\% is not flexible nor suitable. Specifically, the desired integration of a visibility graph and a measure of weather impact-- suggest a more suitable value of visibility is in the original equation for \ac{CR}, \ie using the distance between nodes to determine a contrast ratio rather than a binary threshold. Therefore, for each weather type, the attenuation (also referred to as extinction) coefficient $\sigma$ must be defined (as shown below). The calculations to derive an attenuation coefficient used for this paper are based on Recommendation ITU-R P.1817-1\cite{itu-r}. 

There are two mechanisms involved for the attenuation calculation in the three weather conditions: geometric and aerosol (Mie) scattering. In geometric scattering, the objects of the medium being passed through are large enough that the physical scattering is independent of wavelength. On the other hand, Mie scattering occurs when the particle size is comparable to or slightly smaller than the wavelength of interest. As $\sigma$ is defined for a specific wavelength in Mie scattering, the midrange often used for human-based visibility of $550nm$ is used.

\subsubsection{Rain}\label{sec:rain}

As a rain droplet is much larger than the wavelength of visible light, the function for visibility is wavelength-independent and calculated implicitly. Rain attenuation is defined in~\cite{itu-r} as the power-law:

\begin{equation}\label{eq:rain}
    \sigma_{rain} = k \cdot R^{\alpha} 
\end{equation}
where $k$ is the proportionality parameter, $\alpha$ the proportionality factor, and $R$ the rain rate in $mm/h$. In this paper $k=1.17$, and $\alpha=0.65$, which for validation reproduces the charts in~\cite{itu-r}.

\subsubsection{Snow}\label{sec:snow}

Snow is when water crystallizes, creating large particles and impacting transmission largely through geometric scattering (similar to rain drops). The snow-based calculation of $\sigma$ is defined as:

\begin{equation}
    \sigma_{snow} = a \cdot S^{b} \,\,\texttt{m}^{-1}
\end{equation}
where $S$ is the snowfall rate (mm/h), and $a$, $b$ are functions of wavelength coefficients defined for two types of snow - wet and dry (Table~\ref{tab:snow}) in~\cite{itu-r}.

\begin{table}[!htbp]
\small
\centering
\begin{tabular}{l|cc}
                       & \textbf{a} & \textbf{b}  \\ \hline
Wet    & $0.0001023 \cdot \lambda_{nm}$ +3.7855466  &0.72  \\ \hline
Dry    & $0.0000542 \cdot \lambda_{nm}$ +5.4958776   &1.38                           
\end{tabular}
\caption{Wet and Dry Snow calculation coefficients based on ~\cite{itu-r}. Values are based on $km^{-1}$.}
\label{tab:snow}  
\end{table}

For the dry snow, when applying coefficients from Table \ref{tab:snow}  and with  $\lambda$ = 0.55 ($\mu m$) the scattering coefficient is:

\begin{equation}
    \sigma = 0.0055256876 \cdot S^{1.38} \,\,\texttt{m}^{-1}
\end{equation}
As the $\lambda$ used in this work is within the visible range, it has a negligible impact on results. If visibility was to be considered outside of the visible range (i.e., GHz frequency), $\lambda$ would begin having more considerate impacts. 

\subsubsection{Fog}\label{sec:fog}

Fog is a complex weather phenomenon that is a function of particle sizes and distributions that could encompass both geometric and Mie scattering. Some sources have defined a range of values for $\sigma$ from $0.0015$ to $0.03$~\cite{boyce2008lighting}, which yields visibility at $2\%$ for distances $130$m to $2608.015$m. Others define $\sigma$ for Dense fog between $0.06-0.2$~\cite{folks2000front}. To increase the flexibility and applicability of the analysis methods, instead of relying on pre-determined coefficients for Fog, we implement a physics-based calculation integrating over the particle distributions to determine the Mie scattering contribution.  

The modified gamma distribution for fog particles is defined as:

\begin{equation}
    N(r)=ar^{\alpha} \texttt{exp}(-br)
\end{equation}
The values for the particles used in the modified gamma distribution are seen in Table~\ref{tab:itu-fog}. 

\begin{table}[!htbp]
\small
\begin{tabular}{l|llllll}
                       & \textbf{$\alpha$} & \textbf{$a$} & \textbf{$b$} & \textbf{$N(cm^{-3})$} & \textbf{$W(g/m^3)$} & \textbf{$r_m(\mu m)$}  \\ \hline
HA    & 3                 & 0.027        & 0.3          & 20                   & 0.37                & 10                               \\ \hline
MR & 6                 & 607.5        & 3            & 200                  & 0.02                & 2                            
\end{tabular}
\caption{Attenuation coefficient parameters for Fog in~\cite{itu-r}. (HA) Heavy Advection Fog (MR) Moderate radiation Fog.}
\label{tab:itu-fog}  
\end{table}

The particle size distribution is integrated over absorption (or scattering cross-section) to calculate the attenuation.  The scattering coefficient is defined as:

\begin{equation}
    \sigma_{fog}(\lambda)=10^5 \int \limits_{0}^{\infty} Q_d \left( \frac{2\pi r}{\lambda}, n' \right) \pi r^2 \frac{dN(r)}{dr}dr ~~\text{  km$^{-1}$}
\end{equation}
where $Q_d$ is determined by Mie Scattering theory, $dN(r)/dr$ is particle size distribution per unit of volume (cm${}^{-4}$), $n'$ is the real part of the refractive index $n$ of the aerosol, and $r$ is the radius of the particles (cm)~\cite{hulst1981light}.  Therefore, $\sigma_{fog}$ is found by the distribution function times the scattering cross-section integrated over the radius of the particle sizes.  

At 550(nm) $\sigma = 0.02874311 m^{-1}$ for Heavy Advection Fog. and $\sigma =0.00863808 m^{-1}$ for Moderate Radiation Fog (by dividing the resulting $\sigma$ defined for $km$ by $1000$).

\subsection{Implementation}

The integration of attenuation coefficients with a visibility graph was implemented through a custom C++ package. The package uses Embree~\cite{Wald:2014:EKF:2601097.2601199} raytracing library for highly performant calculations by passing a set of triangles (i.e., from an OBJ file) in 3D. A python package was developed to interface with the C++ code using ctypes and NumPy data structures, which stores distances of ray intersections as data arrays for the $VG$. For each weather condition, an element-wise multiplication is performed using the corresponding $\sigma$ value on the NumPy array. As the code is built to interface through CPython, Section~\ref{sec:results} visualizes many of the basic examples in python, while the complex case is shown in Rhino 3D by using grasshopper to load stored results. While performance is not of focus in this paper, it is important to note the usefulness of fast visibility analysis (e.g., 2D line intersections on GPU~\cite{schneider2012real}).

\subsection{Case Study Models}

\subsubsection{Comparative Models}

\begin{figure}[!htbp]
\centering
\includegraphics[page=1, width=.48\textwidth]{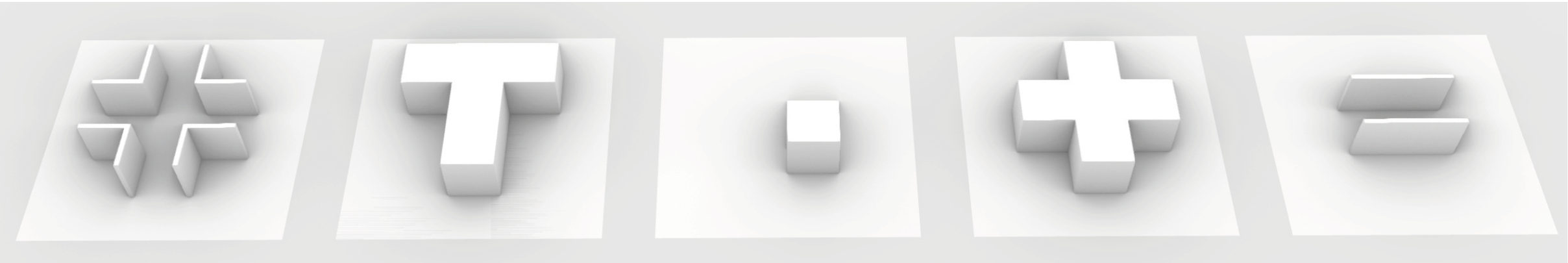}
\caption{3D view of the simple case-studies used in past work to demonstrate visibility graphs.}
\label{fig:vg_cases} 
\end{figure}

To provide a comparison to the breadth of existing research on visibility graphs and space-syntax analysis, a set of basic models that have been seen throughout the literature is used for explaining how the weather-based calculations perform. In Figure~\ref{fig:vg_cases}, five models are shown in a 3D perspective. Each of the models are a 50x50m grid, with extrusion set to 10m. The grid is sampled at 1m spacing, representing a human-scale resolution by which to  divide the space.

\subsubsection{City Generation}

\begin{figure}[!htbp]
\centering
\includegraphics[page=6, width=.48\textwidth]{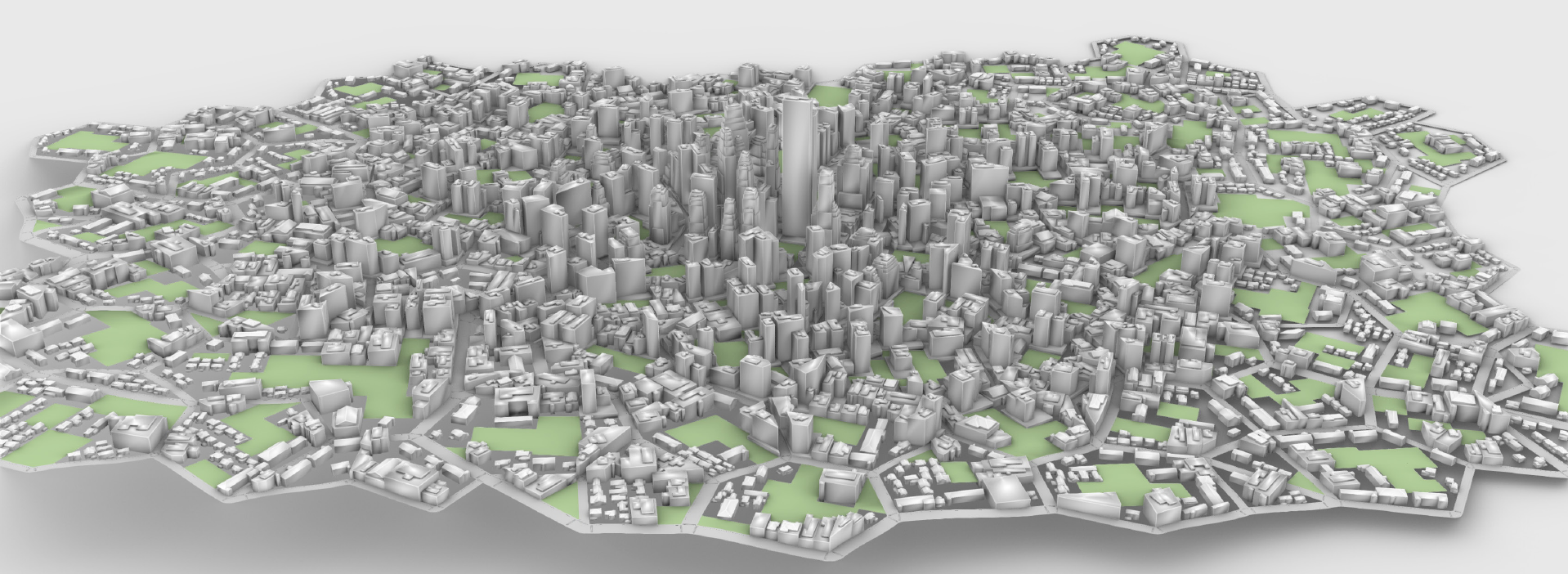}
\caption{3D view of a sample city used for analysis.}
\label{fig:vg_case_city} 
\end{figure}

As a more specific example of when such an approach may be of interest, a sample city was created (Fig.~\ref{fig:vg_case_city}). The environment was created using CityEngine with 2m sidewalks defined around the buildings, using one of the automated features for generating a sample city with a hexagonal street network pattern. The city has varying terrain and building heights, allowing for more complex visibility \ac{LoS} calculations as a hill may occlude nodes otherwise visible in 2D projects.  The sidewalks and crosswalks were used to create a grid of 1m spacing. From the main area of interest, 50,000 nodes were generated to be used in the analysis.

\section{Results}\label{sec:results}

For the following tests, the height to perform \ac{LoS} calculations was set to 1.7 m. The values used for weather conditions are varied, but all within ranges found in the literature.

\subsection{Visibility Graph Case}

\begin{figure}[!htbp]
    \centering
    \begin{subfigure}[t]{\linewidth}
        \includegraphics[width=\textwidth]{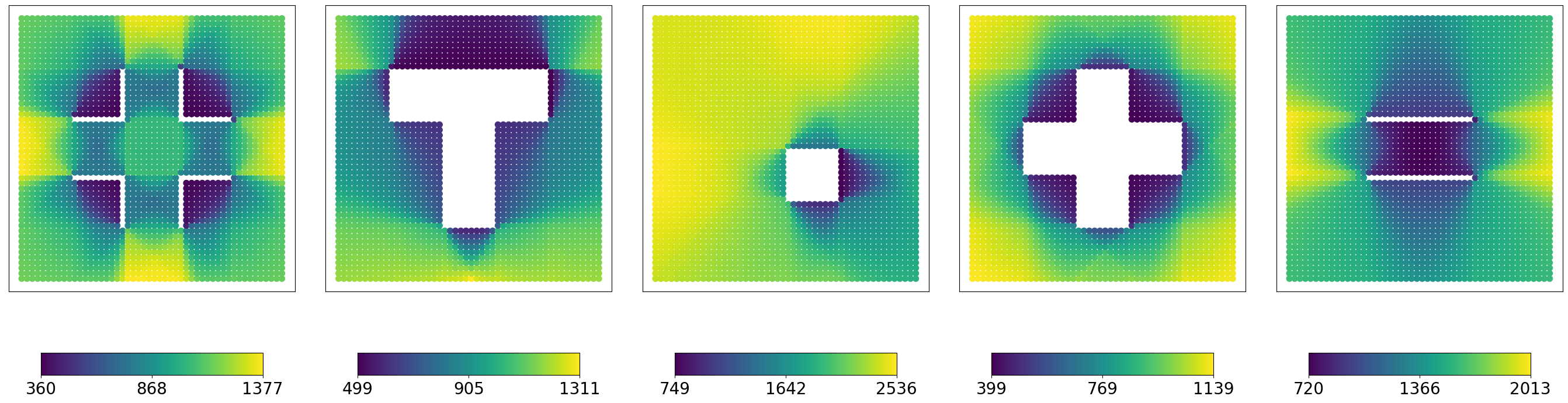}
        \caption{individual}
        \label{subfig:cnt_ind}
    \end{subfigure}
    
    \begin{subfigure}[t]{\linewidth}
        \includegraphics[width=\textwidth]{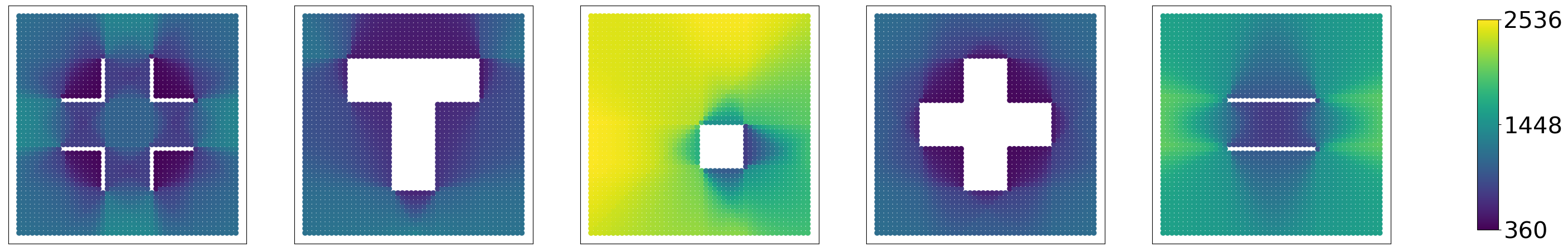}
        \caption{comparison}
        \label{subfig:cnt_all}
    \end{subfigure}
    
    \caption{Degree of each node. In Figure~\ref{subfig:cnt_ind} the color-scale is individual to each case. In Figure~\ref{subfig:cnt_all}, the heatmap of all figures is normalized, showing the third option has the highest number of total edge connections. }
    \label{fig:count}
\end{figure}

First, a basic measure using the models from Figure~\ref{fig:vg_cases} in visibility graph analysis is performed to ensure the system is able to replicate past results. Figure~\ref{fig:count} is the top view of Figure~\ref{fig:vg_cases} shown with the $VG$, where the nodes are colored based on the number of edge connections (\ie number of visible locations), defined as $deg(VG)$. As this result is both logical and follows the visualizations seen in existing works\footnote{\url{https://isovists.org/user_guide/}}, the next step is to understand how \ac{CR} can be analyzed. 

\begin{figure}[!htbp]
\centering
\includegraphics[width=.48\textwidth]{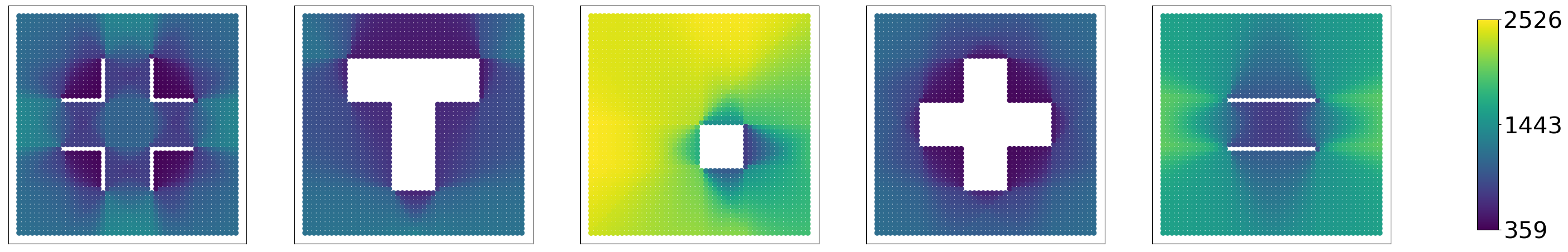}
\caption{Using the summation metric ($S_S$) of edges with clear weather produces indistinguishable results from Fig.~\ref{fig:count}. }
\label{fig:vg_clear} 
\end{figure}

Since clear weather has excellent visibility with minimal impact on \ac{CR} -- especially under 50m (size of the test cases)-- the expectation would be that $deg(v) \approx S_S(v)$, as a \ac{CR} with no extinction on the \ac{LoS} is $1$ (i.e., 100\% contrast). Although in fact, the attenuation is not none, rather $\sigma=0.00015$, the visualization and quantitative results support this expectation, as seen in Figure~\ref{fig:vg_clear}, where the maximum value is 2526 compared to 2536 in Figure~\ref{subfig:cnt_all} and the minimum is different by 1 (359 vs. 360).

As discussed in Sec.~\ref{sec:atten_coefs}, $\sigma$ is a function of the given weather condition, and in the case of rain and snow, a function of precipitation rates. Between these two values, one of which is continuous, there is an unlimited number of tests that can be performed. For brevity, we demonstrate and report on a subset of the possibilities. 

The \ac{CR} is calculated by attenuation, and in many cases of extreme weather, the attenuation coefficient becomes very similar, not providing additional insights into the results.  Therefore, a more modest value of $8(mm/h)$, the upper limit of \textit{Heavy rain}~\cite{usgsRain} is used. For context, adverse weather for rain could indicate a rainfall rate around $50(mm/h)$ as it is a severe rainstorm yet under the maximum rain rate of hurricanes in Florida from the table in~\cite{black2012rain}. For fog, the parameters of Heavy advection (HA in Table~\ref{tab:itu-fog}) are used. Lastly, the snow weather condition needs an input of both type (dry or wet) and rate. Dry snowfall impacts visibility more than wet snow at the same snowfall rate; however, dry snowfall occurs less frequently at higher rates. To illustrate a more impactful weather condition, the calculations for snow use a dry snowfall rate of 4(mm/h), which is high, but within recorded ranges for the type~\cite{rasmussen1999estimation}.

\begin{figure}[!htbp]
    \centering
    \begin{subfigure}[t]{\linewidth}
        \includegraphics[width=\textwidth]{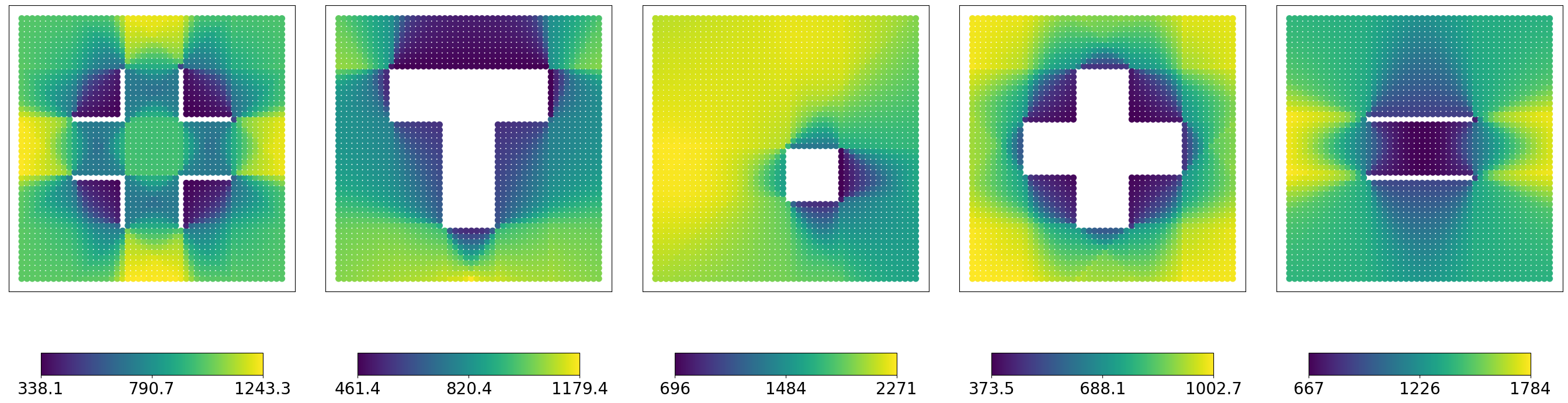}
        \caption{rain}
        \label{subfig:sum_rain}
    \end{subfigure}
    
    \begin{subfigure}[t]{\linewidth}
        \includegraphics[width=\textwidth]{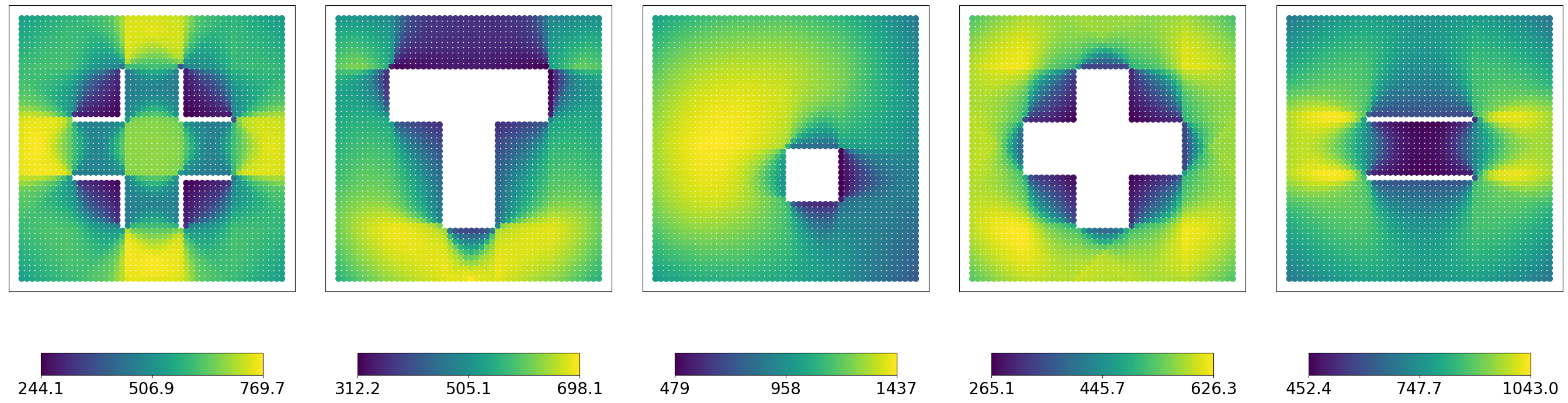}
        \caption{fog}
        \label{subfig:sum_fog}
    \end{subfigure}
    
    \begin{subfigure}[t]{\linewidth}
        \includegraphics[width=\textwidth]{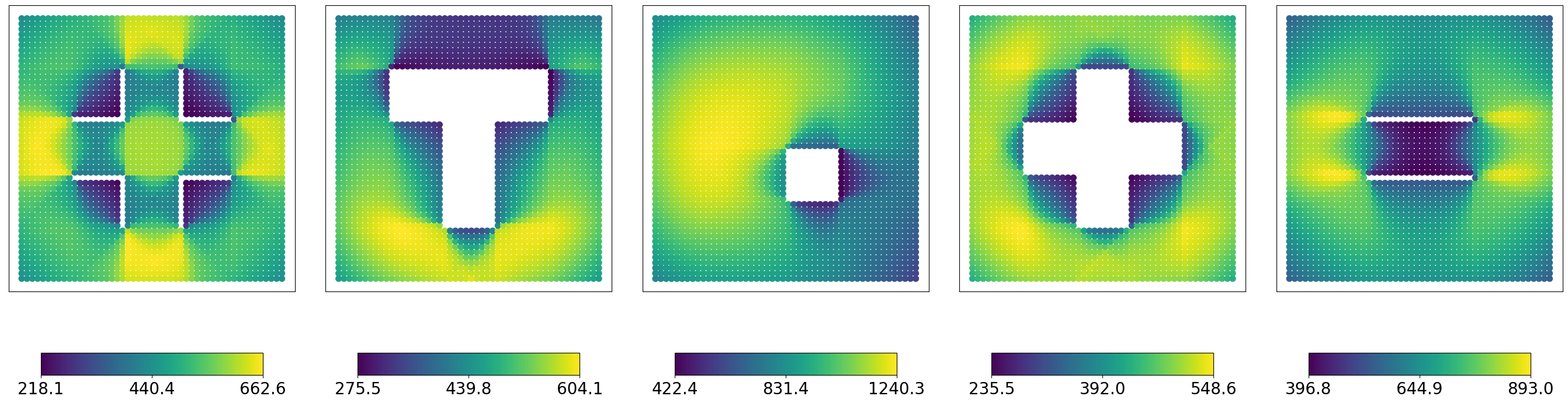}
        \caption{snow}
        \label{subfig:sum_snow}
    \end{subfigure}
    
    \caption{Figure~\ref{subfig:sum_rain} shows results of Rain condition with a rainfall rate of 8(mm/h). Figure~\ref{subfig:sum_fog} shows Heavy Advection fog with parameters from Table~\ref{tab:itu-fog}. Figure~\ref{subfig:sum_snow} shows results of Dry Snow (Table~\ref{tab:snow}) at a rate of 4(mm/h).}
    \label{fig:vg_weather}
\end{figure}

Figure~\ref{fig:vg_weather} shows the impact of each example weather condition on the $VG$ using the $S_S$ metric. The attenuation coefficients with the above-described settings are; Fig.~\ref{subfig:sum_rain} $\sigma=0.0045$, Fig.~\ref{subfig:sum_fog} $\sigma=0.0287$, Fig.~\ref{subfig:sum_snow} $\sigma=0.0374$. Notably, the impact of $\sigma$ on the \ac{CR} applied to the graph attributes is non-linear. While the range of values when moving from the clear to weather conditions can be nearly halved, the heatmap shows it is not a 1-1 correlation. For example, the middle case (single box) in Figure~\ref{subfig:sum_rain} has a $S_S$ range of 696-2271, while the same case in Figure~\ref{subfig:sum_snow} has a range of 411-1240. The first learning from this comparison is the reduction in the score at the high values. The second and more insightful result is in the visualization, showing how the addition of $\sigma$ changes the overall relationship of visible regions. If using only the $deg(v)$ as a method to understand highly visible regions in a design, the basic condition would suggest the mid-left and mid-top areas of this case are ideal. However, once attenuation is accounted for, it becomes apparent that the slight left of the box itself is more visible. 

Of equal interest in these results is the case studies that seem less affected. While each example has a clear reduction in the scores, some (\ie first and last cases) maintain the same areas of high visibility, compared to the third and fourth cases. While by studying these examples, it is possible to understand why certain regions are impacted, it is only possible through using such simplistic models. In a complex city landscape, such associations are impossible without this computational intervention. This would suggest, perhaps, there are design solutions that optimize visibility for both cases (clear and adverse weather), while some designs may require additional considerations. 

In addition to the graphs all-to-all $VG$, another analysis is done with the graph nodes connected to one location. In this example, the general graph of the discretized environment is used as ${}^{a}VG$, and the bottom left corner of the square is used as ${}^{b}VG$. 

\begin{figure}[!htbp]
    \centering
    \begin{subfigure}[t]{0.25\linewidth}
        \includegraphics[trim=0 0 1600 0,clip, width=\textwidth]{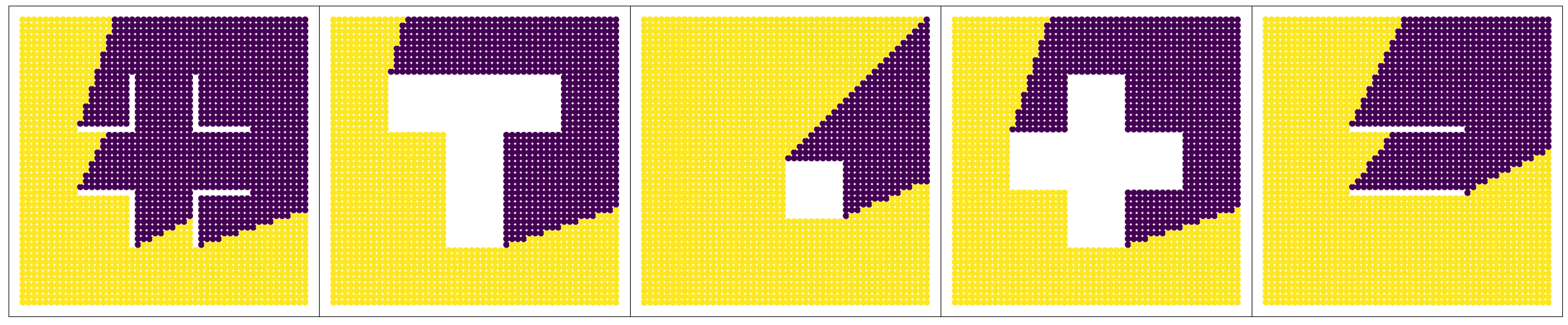}
        \caption{clear}
        \label{subfig:vg-ab_clear}
    \end{subfigure}
    \begin{subfigure}[t]{0.25\linewidth}
        \includegraphics[trim=0 0 1600 0,clip, width=\textwidth]{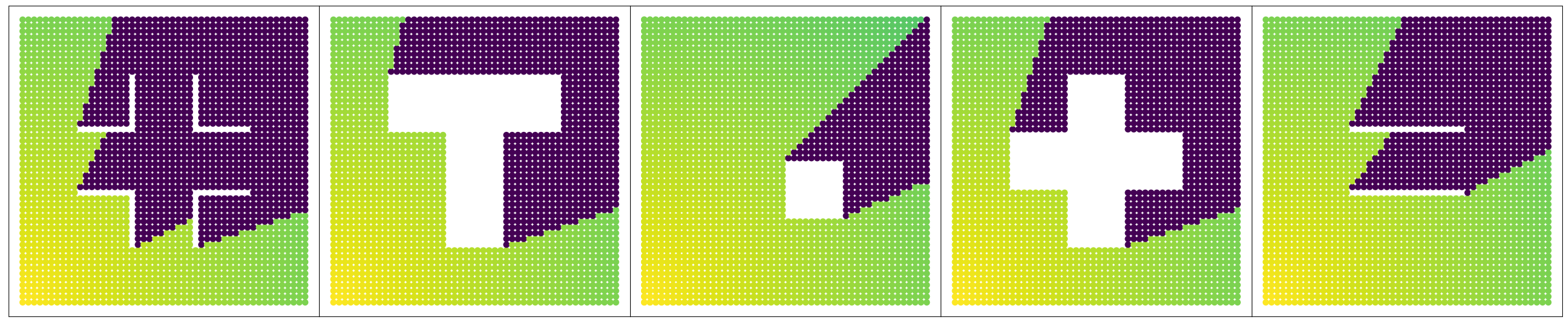}
        \caption{rain}
        \label{subfig:vg-ab_rain}
    \end{subfigure}
    \begin{subfigure}[t]{0.25\linewidth}
        \includegraphics[trim=0 0 1600 0,clip,width=\textwidth]{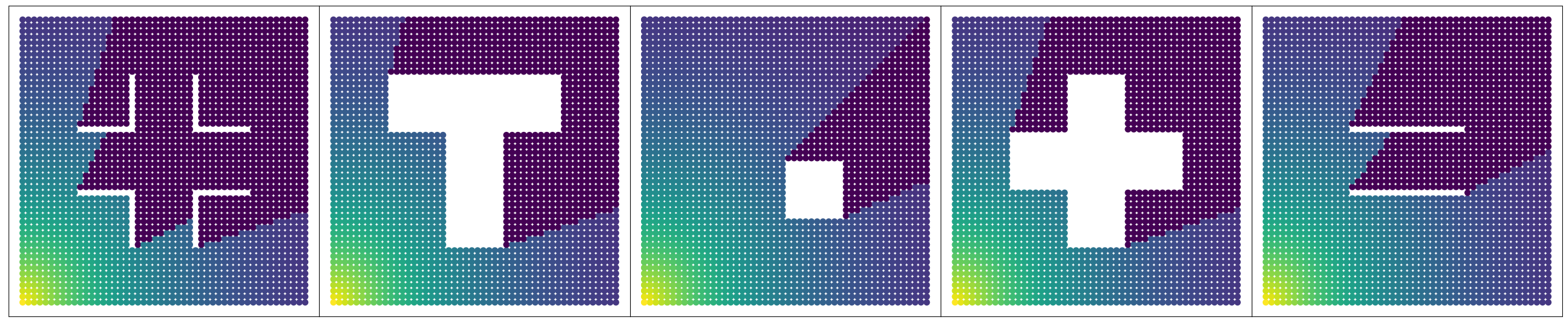}
        \subcaption{snow}
        \label{subfig:vg-ab_snow}
    \end{subfigure}
    
    \caption{ Figs.~\ref{subfig:vg-ab_clear}-\ref{subfig:vg-ab_snow} are shown with a heatmap from 0 (dark purple) to 1 (yellow). }
    \label{fig:vg_ab}
\end{figure}

The results of using the single-point analysis are shown in Figure~\ref{fig:vg_ab}. Like previous results, using the same analysis method and $\sigma=0.00015$ for clear weather condition produced a nearly binary result (Fig.~\ref{subfig:vg-ab_clear}) for visibility of every node in the graph to the bottom left node. The results of using rain (Fig.~\ref{subfig:vg-ab_rain}) and snow (Fig.~\ref{subfig:vg-ab_snow}) begin to illustrate the direct impact on decisions this method can have on design. By selecting a location of interest, the amount of space visible to this location in differing weather conditions (not only in perfect visibility) can help guide decisions such as sign placement. As \ac{CR} is a contrast \textit{ratio}, all the values provided from [0,1] can be multiplied by 100 to convert to a \textit{contrast percentage}. In other words, a \ac{CR} of 0.5 (green in Fig.~\ref{fig:vg_ab}) means the view of that node has 50\% less contrast than in a clear environment. While a black text on a white sign may still be readable, text, signs, or objects that do not \textit{begin} with maximum contrast will be more difficult to see. 

\subsection{Large-scale Case}

\begin{figure}[!htbp]
\centering
\includegraphics[page=2, width=.48\textwidth]{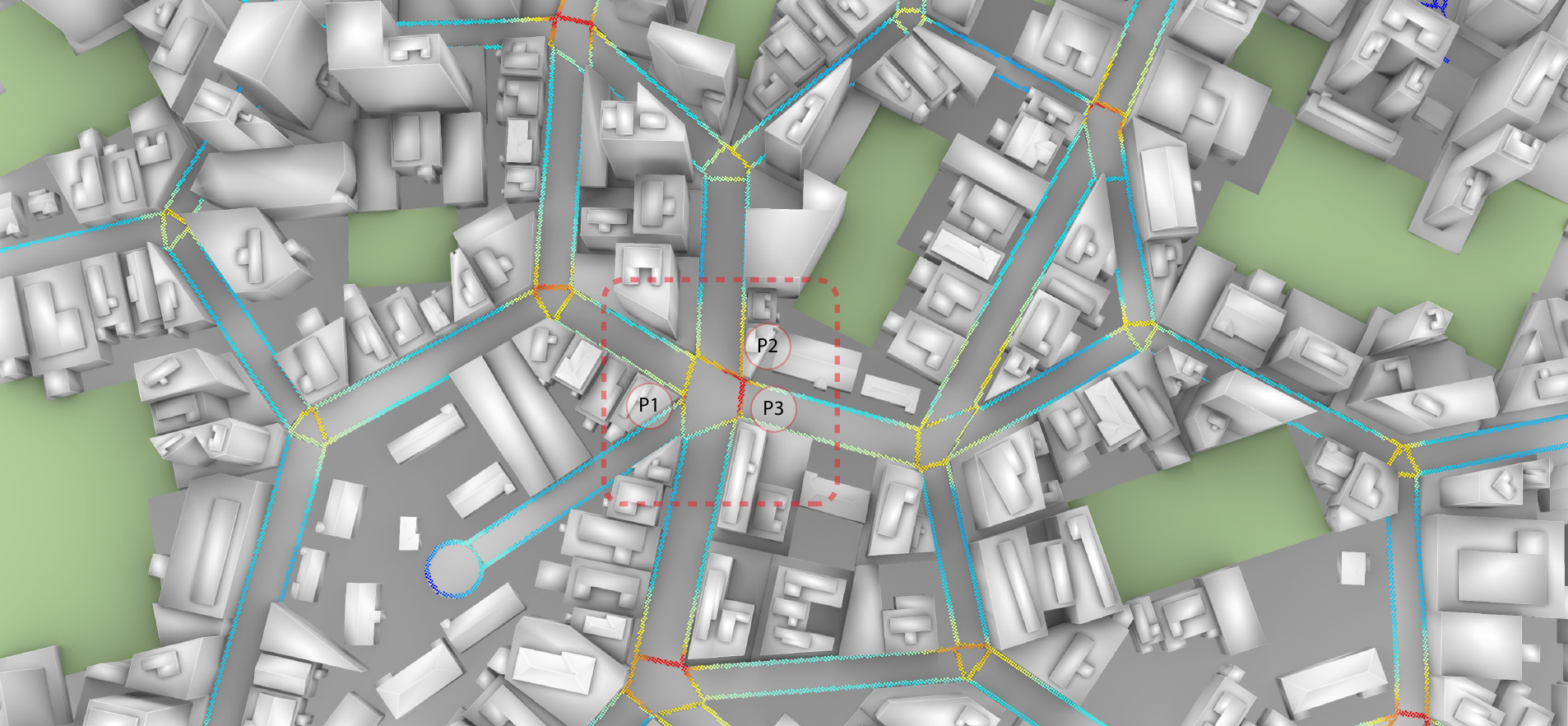}
\caption{Top view of the city with a heatmap visualizing an all-all $VG$ using clear weather. Three points (P1,P2,P3) are identified as areas of interest.  Each are offset as to not obstruct the view of the actual point.  P1 and P2 are both on the corners of the sidewalk and P3 is in the middle of the crosswalk. Color-scale is from high (red) to low (blue) visibility.}
\label{fig:vg_city_region} 
\end{figure}

To demonstrate how the metric may perform when analyzing an urban environment, we focus on a central region of the city model (Fig.~\ref{fig:vg_city_region}). Three sample points are queried at a multi-segment intersection, referred to as P1, P2, and P3. P1 is placed on the sidewalk corner opposite where the $VG$ visually has the highest connectivity. P2 is placed on the corner of high visibility, and P3 is placed on the crosswalk. The crosswalk has the highest visibility as it is not occluded by nearby walls, and is used as an example of how locating signage in the most visible location by \ac{LoS} may not be the most visible in adverse weather. For each point, $S_S$ and $S_A$ were calculated for the clear weather condition; $S_S(P1)=3642$, $S_A(P1)=0.0725$, $S_S(P2)=3664$, $S_A(P2)=0.0730$, $S_S(P3)=4067$, $S_A(P3)=0.0810$. 

\begin{figure}[!htbp]
    \centering
    \begin{subfigure}[t]{.3\linewidth}
        \includegraphics[page=8, width=\textwidth]{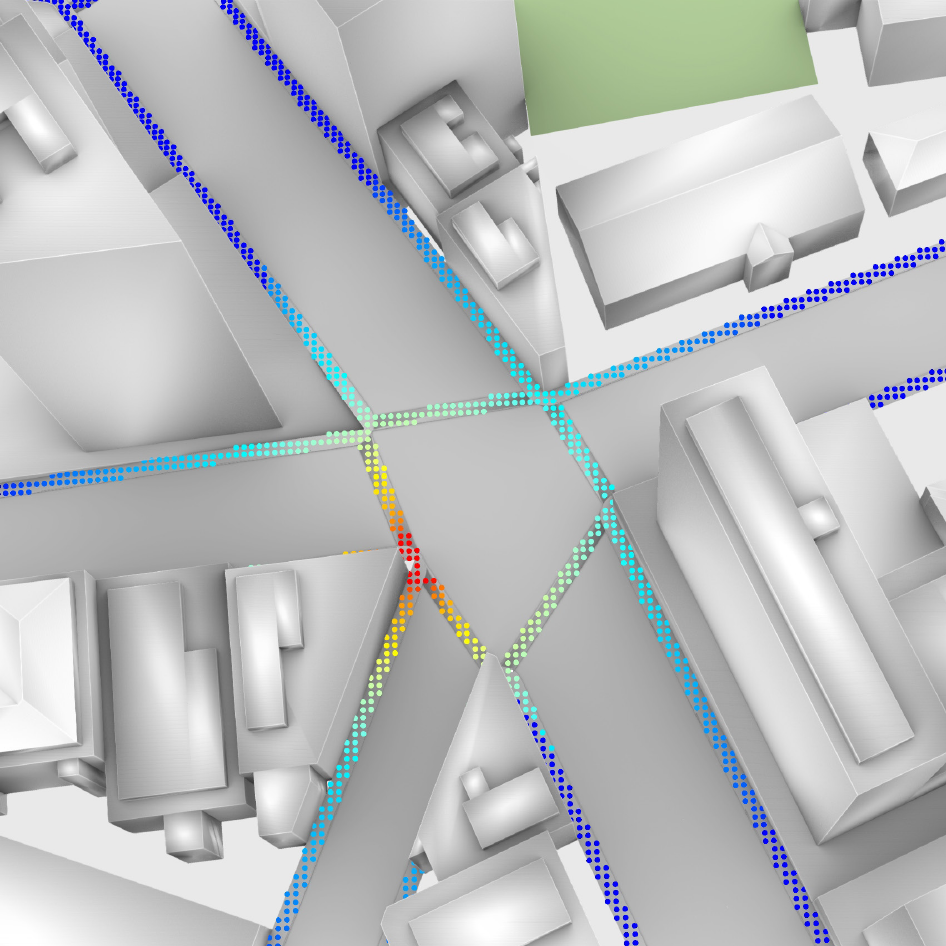}
        \caption{P1}
        \label{subfig:vg-city-p1}
    \end{subfigure}
    \begin{subfigure}[t]{.3\linewidth}
        \includegraphics[page=9, width=\textwidth]{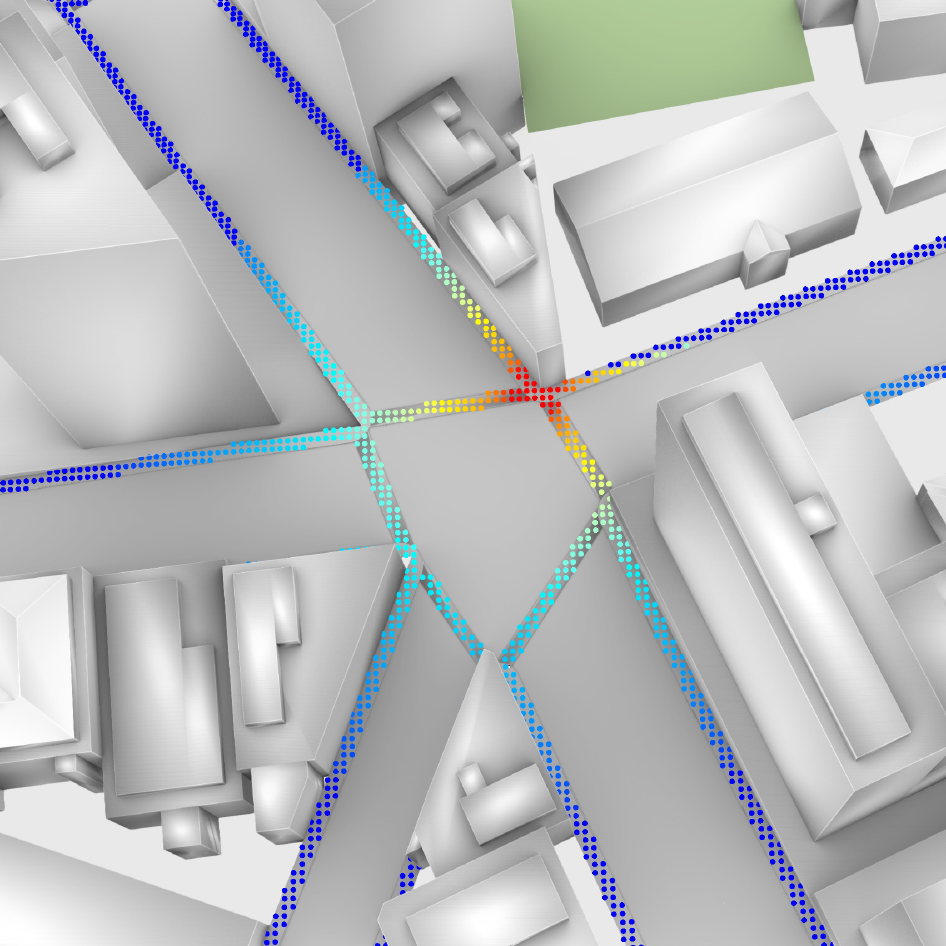}
        \caption{P2}
        \label{subfig:vg-city-p2}
    \end{subfigure}
    \begin{subfigure}[t]{.3\linewidth}
        \includegraphics[page=10, width=\textwidth]{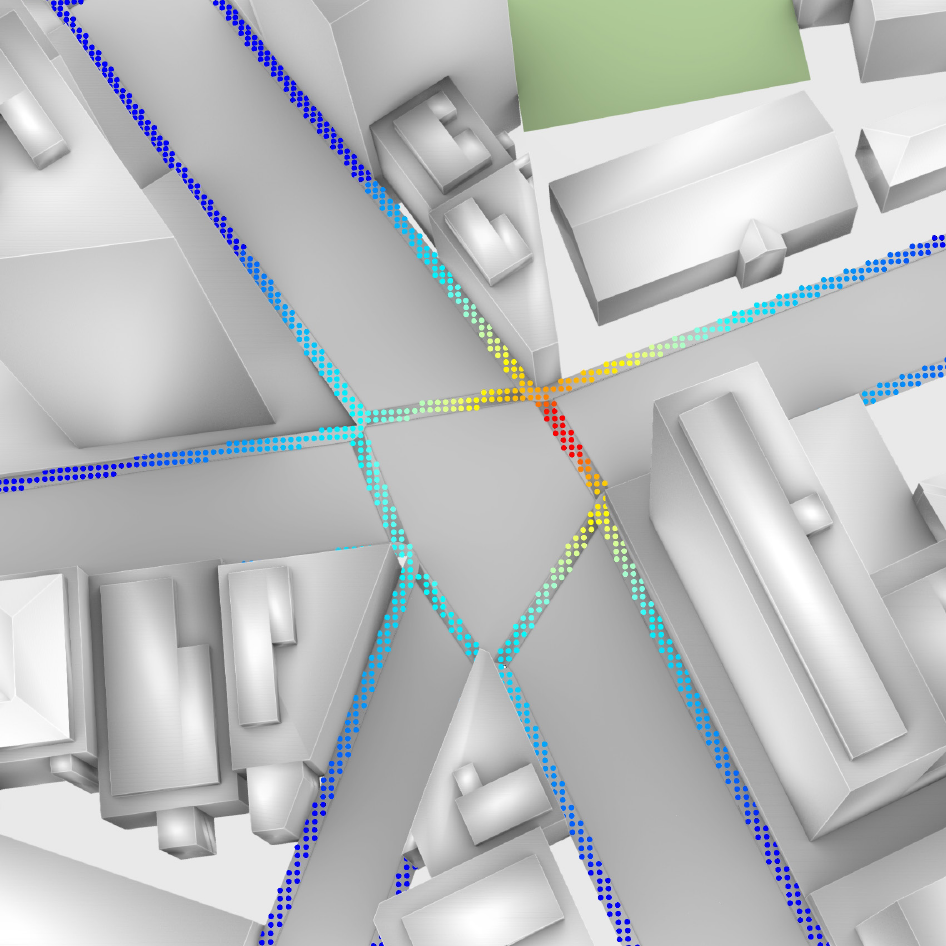}
        \caption{P3}
        \label{subfig:vg-city-p3}
    \end{subfigure}
    \caption{ Heatmap visualizing the $VG$ score for a snowfall condition where the entire $VG$ is ${}^{a}VG$ and for each, case the point is ${}^{b}VG$. Colorscale is from high (red) to low (blue) visibility. Calculated results are: $S_S(P1)=709$, $S_A(P1)=0.014$, $S_S(P2)=610$, $S_A(P2)=0.0121$, $S_S(P3)=686$, $S_A(P3)=0.0136$.}
    \label{fig:vg_city-snow}
\end{figure}

In Figure~\ref{fig:vg_city-snow}, heatmaps showing the visibility of a single node to all others using dry snow at a snowfall rate of 4(mm/h) illustrate the falloff effect of visibility in that snow weather condition. Numerically, while the highest $S_S$ value was P3, factoring in snowfall reduces this value significantly ($S_S(P3)=686$). However, P1, the location \textit{across} from the original high visibility visualization in $VG$, has a value of $S_S(P1)=709$, meaning the relative visibility of P1 is, in fact, better than P3 when accounting for heavy snow.

\section{Conclusions}
The focus of the work is in illustrating how the addition of attenuation coefficients impact existing modes of visualizing and reporting visibility graphs, with special emphasis placed on weather conditions. New modes of visibility-related design can be considered as a result. For example, when considering the limited visibility with large spaces between buildings and roads, modifications to the environment can be made to ensure occupants have visible wayfinding cues and landmarks available to them. The method presented of using extinction coefficients can also be applied to non-weather-related metrics that capture visibility of humans, such as text legibility, similar to the discussion in Sec.~\ref{subsec:physio} (\eg using Snellen test).

During adverse weather conditions for visibility, such as fog, the relationship between seeing far and close is ambiguous. While an occupant may only be able to see a few meters in front of themselves, it is not necessarily the built environment that restricts this distance. Importantly, it is not the built environment that the person sees either. Using a metric that assumes the number of nodes or isovists that are visible as a measure of openness may be problematic in these instances. 


One strength of the work presented is the ease of use and application within existing workflows. As a visibility graph contains the distance between nodes, we can easily retrieve this distance to calculate the  the theoretical contrast ratio. The present work does not include the impact of varying light sources, which changes the contrast ratio through scattering(\ie our weather-based calculations assume a homogeneously illuminated environment). Likewise, time of day (\ie sun angle) is not considered in this work. Future work will consider light placement and distributions with the direction of the visibility graph edge. Likewise, future studies will include analyzing the correlation between reduced visibility and human experience through human-subject studies to understand the meaning (if any) of other space-syntax metrics with this modification.

\begin{acks}
This paper contains work that was supported, or in part by, the U. S. Army Combat Capabilities Development Command (CCDC) Armaments Center and the U. S. Army ManTech Office under Contract Delivery Order W15QKN19F0002 - Advanced Development of Asset Protection Technologies (ADAPT).

The authors thank Anuradha Kadam for early work on this concept, Krunali Shah for work on the diagrams, and Drew Balletto for assistance on programming.
\end{acks}

\bibliographystyle{ACM-Reference-Format}
\bibliography{visibility_ref}


\begin{thebibliography}{40}


\ifx \showCODEN    \undefined \def \showCODEN     #1{\unskip}     \fi
\ifx \showDOI      \undefined \def \showDOI       #1{#1}\fi
\ifx \showISBNx    \undefined \def \showISBNx     #1{\unskip}     \fi
\ifx \showISBNxiii \undefined \def \showISBNxiii  #1{\unskip}     \fi
\ifx \showISSN     \undefined \def \showISSN      #1{\unskip}     \fi
\ifx \showLCCN     \undefined \def \showLCCN      #1{\unskip}     \fi
\ifx \shownote     \undefined \def \shownote      #1{#1}          \fi
\ifx \showarticletitle \undefined \def \showarticletitle #1{#1}   \fi
\ifx \showURL      \undefined \def \showURL       {\relax}        \fi
\providecommand\bibfield[2]{#2}
\providecommand\bibinfo[2]{#2}
\providecommand\natexlab[1]{#1}
\providecommand\showeprint[2][]{arXiv:#2}

\bibitem[\protect\citeauthoryear{Benedikt}{Benedikt}{1979}]%
        {benedikt1979take}
\bibfield{author}{\bibinfo{person}{Michael~L Benedikt}.}
  \bibinfo{year}{1979}\natexlab{}.
\newblock \showarticletitle{To take hold of space: isovists and isovist
  fields}.
\newblock \bibinfo{journal}{\emph{Environment and Planning B: Planning and
  design}} \bibinfo{volume}{6}, \bibinfo{number}{1} (\bibinfo{year}{1979}),
  \bibinfo{pages}{47--65}.
\newblock


\bibitem[\protect\citeauthoryear{Black and Hallett}{Black and Hallett}{2012}]%
        {black2012rain}
\bibfield{author}{\bibinfo{person}{Robert~A Black} {and} \bibinfo{person}{John
  Hallett}.} \bibinfo{year}{2012}\natexlab{}.
\newblock \showarticletitle{Rain rate and water content in hurricanes compared
  with summer rain in Miami, Florida}.
\newblock \bibinfo{journal}{\emph{Journal of Applied Meteorology and
  Climatology}} \bibinfo{volume}{51}, \bibinfo{number}{12}
  (\bibinfo{year}{2012}), \bibinfo{pages}{2218--2235}.
\newblock


\bibitem[\protect\citeauthoryear{Boyce}{Boyce}{2008}]%
        {boyce2008lighting}
\bibfield{author}{\bibinfo{person}{Peter~R Boyce}.}
  \bibinfo{year}{2008}\natexlab{}.
\newblock \bibinfo{booktitle}{\emph{Lighting for driving: roads, vehicles,
  signs, and signals}}.
\newblock \bibinfo{publisher}{CRC Press}.
\newblock


\bibitem[\protect\citeauthoryear{Choi}{Choi}{1999}]%
        {choi1999morphology}
\bibfield{author}{\bibinfo{person}{Yoon~Kyung Choi}.}
  \bibinfo{year}{1999}\natexlab{}.
\newblock \showarticletitle{The morphology of exploration and encounter in
  museum layouts}.
\newblock \bibinfo{journal}{\emph{Environment and Planning B: Planning and
  Design}} \bibinfo{volume}{26}, \bibinfo{number}{2} (\bibinfo{year}{1999}),
  \bibinfo{pages}{241--250}.
\newblock


\bibitem[\protect\citeauthoryear{Conroy}{Conroy}{2001}]%
        {conroy2001spatial}
\bibfield{author}{\bibinfo{person}{Ruth~Alison Conroy}.}
  \bibinfo{year}{2001}\natexlab{}.
\newblock \emph{\bibinfo{title}{Spatial navigation in immersive virtual
  environments}}.
\newblock \bibinfo{thesistype}{Ph.D. Dissertation}. \bibinfo{school}{Citeseer}.
\newblock


\bibitem[\protect\citeauthoryear{Fischer and Knutti}{Fischer and
  Knutti}{2016}]%
        {fischer2016observed}
\bibfield{author}{\bibinfo{person}{Erich~M Fischer} {and} \bibinfo{person}{Reto
  Knutti}.} \bibinfo{year}{2016}\natexlab{}.
\newblock \showarticletitle{Observed heavy precipitation increase confirms
  theory and early models}.
\newblock \bibinfo{journal}{\emph{Nature Climate Change}} \bibinfo{volume}{6},
  \bibinfo{number}{11} (\bibinfo{year}{2016}), \bibinfo{pages}{986--991}.
\newblock


\bibitem[\protect\citeauthoryear{Fisher-Gewirtzman}{Fisher-Gewirtzman}{2014}]%
        {fisher20143d}
\bibfield{author}{\bibinfo{person}{Dafna Fisher-Gewirtzman}.}
  \bibinfo{year}{2014}\natexlab{}.
\newblock \showarticletitle{3D LOS visibility analysis model: Incorporating
  quantitative/qualitative aspects in urban environments}.
\newblock In \bibinfo{booktitle}{\emph{Geodesign by Integrating Design and
  Geospatial Sciences}}. \bibinfo{publisher}{Springer},
  \bibinfo{pages}{219--236}.
\newblock


\bibitem[\protect\citeauthoryear{Fisher-Gewirtzman}{Fisher-Gewirtzman}{2017}]%
        {fisher2017can}
\bibfield{author}{\bibinfo{person}{DAFNA Fisher-Gewirtzman}.}
  \bibinfo{year}{2017}\natexlab{}.
\newblock \showarticletitle{Can 3D Visibility Calculations along a Path Predict
  the Perceived Density of Participants Immersed in a Virtual Reality
  Environment?}. In \bibinfo{booktitle}{\emph{Proceedings of the Eleventh
  International Space Syntax Symposium}}. \bibinfo{pages}{160--1}.
\newblock


\bibitem[\protect\citeauthoryear{Fisher-Gewirtzman}{Fisher-Gewirtzman}{2018}]%
        {fisher2018integrating}
\bibfield{author}{\bibinfo{person}{Dafna Fisher-Gewirtzman}.}
  \bibinfo{year}{2018}\natexlab{}.
\newblock \showarticletitle{Integrating ‘weighted views’ to quantitative 3D
  visibility analysis as a predictive tool for perception of space}.
\newblock \bibinfo{journal}{\emph{Environment and Planning B: Urban Analytics
  and City Science}} \bibinfo{volume}{45}, \bibinfo{number}{2}
  (\bibinfo{year}{2018}), \bibinfo{pages}{345--366}.
\newblock


\bibitem[\protect\citeauthoryear{Folks and Kreysar}{Folks and Kreysar}{2000}]%
        {folks2000front}
\bibfield{author}{\bibinfo{person}{William~R Folks} {and}
  \bibinfo{person}{Douglas Kreysar}.} \bibinfo{year}{2000}\natexlab{}.
\newblock \bibinfo{booktitle}{\emph{Front fog lamp performance}}.
\newblock \bibinfo{type}{{T}echnical {R}eport}. \bibinfo{institution}{SAE
  Technical Paper}.
\newblock


\bibitem[\protect\citeauthoryear{Fuchkina}{Fuchkina}{2017}]%
        {Fuchkina2017}
\bibfield{author}{\bibinfo{person}{Ekaterina Fuchkina}.}
  \bibinfo{year}{2017}\natexlab{}.
\newblock \showarticletitle{Pedestrian Movement Graph Analysis}.
\newblock \bibinfo{journal}{\emph{Arbeitspapiere Informatik in der
  Architektur}} (\bibinfo{year}{2017}).
\newblock
\urldef\tempurl%
\url{https://doi.org/10.25643/bauhaus-universitaet.2738}
\showDOI{\tempurl}


\bibitem[\protect\citeauthoryear{Gordon}{Gordon}{1979}]%
        {gordon1979daytime}
\bibfield{author}{\bibinfo{person}{Jacqueline~I Gordon}.}
  \bibinfo{year}{1979}\natexlab{}.
\newblock \bibinfo{booktitle}{\emph{Daytime visibility, a conceptual review}}.
\newblock \bibinfo{type}{{T}echnical {R}eport}. \bibinfo{institution}{SCRIPPS
  INSTITUTION OF OCEANOGRAPHY LA JOLLA CA VISABILITY LAB}.
\newblock


\bibitem[\protect\citeauthoryear{Greenberg, Natapov, and
  Fisher-Gewirtzman}{Greenberg et~al\mbox{.}}{2020}]%
        {greenberg2020physical}
\bibfield{author}{\bibinfo{person}{Eliyahu Greenberg}, \bibinfo{person}{Asya
  Natapov}, {and} \bibinfo{person}{Dafna Fisher-Gewirtzman}.}
  \bibinfo{year}{2020}\natexlab{}.
\newblock \showarticletitle{A physical effort-based model for pedestrian
  movement in topographic urban environments}.
\newblock \bibinfo{journal}{\emph{Journal of Urban Design}}
  \bibinfo{volume}{25}, \bibinfo{number}{1} (\bibinfo{year}{2020}),
  \bibinfo{pages}{86--107}.
\newblock


\bibitem[\protect\citeauthoryear{Hulst and van~de Hulst}{Hulst and van~de
  Hulst}{1981}]%
        {hulst1981light}
\bibfield{author}{\bibinfo{person}{Hendrik~Christoffel Hulst} {and}
  \bibinfo{person}{Hendrik~C van~de Hulst}.} \bibinfo{year}{1981}\natexlab{}.
\newblock \bibinfo{booktitle}{\emph{Light scattering by small particles}}.
\newblock \bibinfo{publisher}{Courier Corporation}.
\newblock


\bibitem[\protect\citeauthoryear{Kallmann and Kapadia}{Kallmann and
  Kapadia}{2014}]%
        {kallmann2014navigation}
\bibfield{author}{\bibinfo{person}{Marcelo Kallmann} {and}
  \bibinfo{person}{Mubbasir Kapadia}.} \bibinfo{year}{2014}\natexlab{}.
\newblock \showarticletitle{Navigation meshes and real-time dynamic planning
  for virtual worlds}. In \bibinfo{booktitle}{\emph{ACM SIGGRAPH 2014
  Courses}}. ACM, \bibinfo{pages}{3}.
\newblock


\bibitem[\protect\citeauthoryear{Koltsova, Tun{\c{c}}er, and Schmitt}{Koltsova
  et~al\mbox{.}}{2013}]%
        {koltsova2013visibility}
\bibfield{author}{\bibinfo{person}{Anastasia Koltsova}, \bibinfo{person}{Bige
  Tun{\c{c}}er}, {and} \bibinfo{person}{Gerhard Schmitt}.}
  \bibinfo{year}{2013}\natexlab{}.
\newblock \showarticletitle{Visibility analysis for 3D urban environments}.
\newblock  (\bibinfo{year}{2013}).
\newblock


\bibitem[\protect\citeauthoryear{Kruse, McGlauchlin, and McQuistan}{Kruse
  et~al\mbox{.}}{1962}]%
        {kruse1962elements}
\bibfield{author}{\bibinfo{person}{Paul~W Kruse}, \bibinfo{person}{Laurence~D
  McGlauchlin}, {and} \bibinfo{person}{Richmond~B McQuistan}.}
  \bibinfo{year}{1962}\natexlab{}.
\newblock \showarticletitle{Elements of infrared technology: Generation,
  transmission and detection}.
\newblock \bibinfo{journal}{\emph{New York: Wiley, 1962}}
  (\bibinfo{year}{1962}).
\newblock


\bibitem[\protect\citeauthoryear{Lamarche}{Lamarche}{2009}]%
        {lamarche2009topoplan}
\bibfield{author}{\bibinfo{person}{Fabrice Lamarche}.}
  \bibinfo{year}{2009}\natexlab{}.
\newblock \showarticletitle{Topoplan: a topological path planner for real time
  human navigation under floor and ceiling constraints}. In
  \bibinfo{booktitle}{\emph{Computer Graphics Forum}},
  Vol.~\bibinfo{volume}{28}. Wiley Online Library, \bibinfo{pages}{649--658}.
\newblock


\bibitem[\protect\citeauthoryear{Lee, Eastman, Lee, Kannala, and Jeong}{Lee
  et~al\mbox{.}}{2010}]%
        {lee2010computing}
\bibfield{author}{\bibinfo{person}{Jin-kook Lee}, \bibinfo{person}{Charles~M
  Eastman}, \bibinfo{person}{Jaemin Lee}, \bibinfo{person}{Matti Kannala},
  {and} \bibinfo{person}{Yeon-suk Jeong}.} \bibinfo{year}{2010}\natexlab{}.
\newblock \showarticletitle{Computing walking distances within buildings using
  the universal circulation network}.
\newblock \bibinfo{journal}{\emph{Environment and Planning B: Planning and
  Design}} \bibinfo{volume}{37}, \bibinfo{number}{4} (\bibinfo{year}{2010}),
  \bibinfo{pages}{628--645}.
\newblock


\bibitem[\protect\citeauthoryear{Lee and Shang}{Lee and Shang}{2016}]%
        {lee2016visibility}
\bibfield{author}{\bibinfo{person}{Zhongping Lee} {and}
  \bibinfo{person}{Shaoling Shang}.} \bibinfo{year}{2016}\natexlab{}.
\newblock \showarticletitle{Visibility: How applicable is the century-old
  Koschmieder model?}
\newblock \bibinfo{journal}{\emph{Journal of the Atmospheric Sciences}}
  \bibinfo{volume}{73}, \bibinfo{number}{11} (\bibinfo{year}{2016}),
  \bibinfo{pages}{4573--4581}.
\newblock


\bibitem[\protect\citeauthoryear{Lozano-P{\'e}rez and Wesley}{Lozano-P{\'e}rez
  and Wesley}{1979}]%
        {lozano1979algorithm}
\bibfield{author}{\bibinfo{person}{Tom{\'a}s Lozano-P{\'e}rez} {and}
  \bibinfo{person}{Michael~A Wesley}.} \bibinfo{year}{1979}\natexlab{}.
\newblock \showarticletitle{An algorithm for planning collision-free paths
  among polyhedral obstacles}.
\newblock \bibinfo{journal}{\emph{Commun. ACM}} \bibinfo{volume}{22},
  \bibinfo{number}{10} (\bibinfo{year}{1979}), \bibinfo{pages}{560--570}.
\newblock


\bibitem[\protect\citeauthoryear{Lu, Gou, Ye, and Sheng}{Lu
  et~al\mbox{.}}{2019}]%
        {lu2019three}
\bibfield{author}{\bibinfo{person}{Yi Lu}, \bibinfo{person}{Zhonghua Gou},
  \bibinfo{person}{Yu Ye}, {and} \bibinfo{person}{Qiang Sheng}.}
  \bibinfo{year}{2019}\natexlab{}.
\newblock \showarticletitle{Three-dimensional visibility graph analysis and its
  application}.
\newblock \bibinfo{journal}{\emph{Environment and Planning B: Urban Analytics
  and City Science}} \bibinfo{volume}{46}, \bibinfo{number}{5}
  (\bibinfo{year}{2019}), \bibinfo{pages}{948--962}.
\newblock


\bibitem[\protect\citeauthoryear{Luebke, Reddy, Cohen, Varshney, Watson, and
  Huebner}{Luebke et~al\mbox{.}}{2003}]%
        {luebke2003level}
\bibfield{author}{\bibinfo{person}{David Luebke}, \bibinfo{person}{Martin
  Reddy}, \bibinfo{person}{Jonathan~D Cohen}, \bibinfo{person}{Amitabh
  Varshney}, \bibinfo{person}{Benjamin Watson}, {and} \bibinfo{person}{Robert
  Huebner}.} \bibinfo{year}{2003}\natexlab{}.
\newblock \bibinfo{booktitle}{\emph{Level of detail for 3D graphics}}.
\newblock \bibinfo{publisher}{Morgan Kaufmann}.
\newblock


\bibitem[\protect\citeauthoryear{Miao, Welch, and Sriraj}{Miao
  et~al\mbox{.}}{2019}]%
        {miao2019extreme}
\bibfield{author}{\bibinfo{person}{Qing Miao}, \bibinfo{person}{Eric~W Welch},
  {and} \bibinfo{person}{PS Sriraj}.} \bibinfo{year}{2019}\natexlab{}.
\newblock \showarticletitle{Extreme weather, public transport ridership and
  moderating effect of bus stop shelters}.
\newblock \bibinfo{journal}{\emph{Journal of Transport Geography}}
  \bibinfo{volume}{74} (\bibinfo{year}{2019}), \bibinfo{pages}{125--133}.
\newblock


\bibitem[\protect\citeauthoryear{Nagy, Villaggi, Stoddart, and Benjamin}{Nagy
  et~al\mbox{.}}{2017}]%
        {nagy2017buzz}
\bibfield{author}{\bibinfo{person}{Danil Nagy}, \bibinfo{person}{Lorenzo
  Villaggi}, \bibinfo{person}{James Stoddart}, {and} \bibinfo{person}{David
  Benjamin}.} \bibinfo{year}{2017}\natexlab{}.
\newblock \showarticletitle{The Buzz Metric: A Graph-based Method for
  Quantifying Productive Congestion in Generative Space Planning for
  Architecture}.
\newblock \bibinfo{journal}{\emph{Technology| Architecture+ Design}}
  \bibinfo{volume}{1}, \bibinfo{number}{2} (\bibinfo{year}{2017}),
  \bibinfo{pages}{186--195}.
\newblock


\bibitem[\protect\citeauthoryear{Nutsford, Reitsma, Pearson, and
  Kingham}{Nutsford et~al\mbox{.}}{2015}]%
        {nutsford2015personalising}
\bibfield{author}{\bibinfo{person}{Daniel Nutsford}, \bibinfo{person}{Femke
  Reitsma}, \bibinfo{person}{Amber~L Pearson}, {and} \bibinfo{person}{Simon
  Kingham}.} \bibinfo{year}{2015}\natexlab{}.
\newblock \showarticletitle{Personalising the viewshed: Visibility analysis
  from the human perspective}.
\newblock \bibinfo{journal}{\emph{Applied Geography}}  \bibinfo{volume}{62}
  (\bibinfo{year}{2015}), \bibinfo{pages}{1--7}.
\newblock


\bibitem[\protect\citeauthoryear{of~ITU}{of~ITU}{[n.d.]}]%
        {itu-r}
\bibfield{author}{\bibinfo{person}{{}Radiocommunication~Sector of ITU}.}
  \bibinfo{year}{[n.d.]}\natexlab{}.
\newblock \bibinfo{booktitle}{\emph{{Recommendation ITU-R P.1817-1} Propagation
  data required for the design of terrestrial free-space optical links}}.
\newblock \bibinfo{type}{{T}echnical {R}eport}.
\newblock


\bibitem[\protect\citeauthoryear{P{\'e}roche, L{\'e}one, and
  Gutton}{P{\'e}roche et~al\mbox{.}}{2014}]%
        {peroche2014accessibility}
\bibfield{author}{\bibinfo{person}{Matthieu P{\'e}roche},
  \bibinfo{person}{Fr{\'e}d{\'e}ric L{\'e}one}, {and} \bibinfo{person}{RJ
  Gutton}.} \bibinfo{year}{2014}\natexlab{}.
\newblock \showarticletitle{An accessibility graph-based model to optimize
  tsunami evacuation sites and routes in Martinique, France}.
\newblock \bibinfo{journal}{\emph{Advances in Geosciences}}
  (\bibinfo{year}{2014}).
\newblock


\bibitem[\protect\citeauthoryear{Pettre, Laumond, and Thalmann}{Pettre
  et~al\mbox{.}}{2005}]%
        {pettre2005navigation}
\bibfield{author}{\bibinfo{person}{Julien Pettre}, \bibinfo{person}{Jean-Paul
  Laumond}, {and} \bibinfo{person}{Daniel Thalmann}.}
  \bibinfo{year}{2005}\natexlab{}.
\newblock \showarticletitle{A navigation graph for real-time crowd animation on
  multilayered and uneven terrain}. In \bibinfo{booktitle}{\emph{First
  International Workshop on Crowd Simulation}}, Vol.~\bibinfo{volume}{43}. New
  York: Pergamon Press, \bibinfo{pages}{194}.
\newblock


\bibitem[\protect\citeauthoryear{Rasmussen, Vivekanandan, Cole, Myers, and
  Masters}{Rasmussen et~al\mbox{.}}{1999}]%
        {rasmussen1999estimation}
\bibfield{author}{\bibinfo{person}{Roy~M Rasmussen}, \bibinfo{person}{Jothiram
  Vivekanandan}, \bibinfo{person}{Jeffrey Cole}, \bibinfo{person}{Barry Myers},
  {and} \bibinfo{person}{Charles Masters}.} \bibinfo{year}{1999}\natexlab{}.
\newblock \showarticletitle{The Estimation of Snowfall Rate Using Visibility}.
\newblock \bibinfo{journal}{\emph{Journal of Applied Meteorology}}
  \bibinfo{volume}{38}, \bibinfo{number}{10} (\bibinfo{year}{1999}),
  \bibinfo{pages}{1542--1563}.
\newblock


\bibitem[\protect\citeauthoryear{Schneider and K{\"o}nig}{Schneider and
  K{\"o}nig}{2012}]%
        {schneider2012real}
\bibfield{author}{\bibinfo{person}{S Schneider} {and} \bibinfo{person}{R
  K{\"o}nig}.} \bibinfo{year}{2012}\natexlab{}.
\newblock \showarticletitle{Real-Time Visibility Analysis-Enhancing calculation
  speed of isovists and isovist-fields using the GPU}. In
  \bibinfo{booktitle}{\emph{12th International Conference on Design \& Decision
  Support Systems in Architecture and Urban Planning, Eindhoven, Netherlands}}.
\newblock


\bibitem[\protect\citeauthoryear{Schwartz}{Schwartz}{2020}]%
        {Schwartz:jogging}
\bibfield{author}{\bibinfo{person}{Mathew Schwartz}.}
  \bibinfo{year}{2020}\natexlab{}.
\newblock \showarticletitle{Evaluating Jogging Routes in Mass Models}. In
  \bibinfo{booktitle}{\emph{Proceedings of the Symposium on Simulation for
  Architecture and Urban Design}} (Online) \emph{(\bibinfo{series}{SIMAUD
  '20})}. \bibinfo{publisher}{Society for Computer Simulation International},
  \bibinfo{address}{San Diego, CA, USA}.
\newblock


\bibitem[\protect\citeauthoryear{Schwartz and Das}{Schwartz and Das}{2019}]%
        {Schwartz:2019:nonflat}
\bibfield{author}{\bibinfo{person}{Mathew Schwartz} {and}
  \bibinfo{person}{Subhajit Das}.} \bibinfo{year}{2019}\natexlab{}.
\newblock \showarticletitle{Interpretting non-flat surfaces for walkability
  analysis}. In \bibinfo{booktitle}{\emph{Proceedings of the Symposium on
  Simulation for Architecture and Urban Design}} (Altanta, Georgia)
  \emph{(\bibinfo{series}{SIMAUD '19})}. \bibinfo{publisher}{Society for
  Computer Simulation International}, \bibinfo{address}{San Diego, CA, USA},
  Article \bibinfo{articleno}{19}, \bibinfo{numpages}{8}~pages.
\newblock
\showISBNx{978-1-5108-6315-6}


\bibitem[\protect\citeauthoryear{Shin and Lee}{Shin and Lee}{2019}]%
        {shin2019indoor}
\bibfield{author}{\bibinfo{person}{Jaeyoung Shin} {and}
  \bibinfo{person}{Jin-Kook Lee}.} \bibinfo{year}{2019}\natexlab{}.
\newblock \showarticletitle{Indoor Walkability Index: BIM-enabled approach to
  Quantifying building circulation}.
\newblock \bibinfo{journal}{\emph{Automation in Construction}}
  \bibinfo{volume}{106} (\bibinfo{year}{2019}), \bibinfo{pages}{102845}.
\newblock


\bibitem[\protect\citeauthoryear{Survey}{Survey}{2020}]%
        {usgsRain}
\bibfield{author}{\bibinfo{person}{United States~Geological Survey}.}
  \bibinfo{year}{2020}\natexlab{}.
\newblock \bibinfo{title}{{Rainfall calculator, metric units How much water
  falls during a storm}}.
\newblock
  \bibinfo{howpublished}{\url{https://water.usgs.gov/edu/activity-howmuchrain-metric.html}}.
\newblock


\bibitem[\protect\citeauthoryear{Suter}{Suter}{2013}]%
        {suter2013structure}
\bibfield{author}{\bibinfo{person}{Georg Suter}.}
  \bibinfo{year}{2013}\natexlab{}.
\newblock \showarticletitle{Structure and spatial consistency of network-based
  space layouts for building and product design}.
\newblock \bibinfo{journal}{\emph{Computer-Aided Design}} \bibinfo{volume}{45},
  \bibinfo{number}{8-9} (\bibinfo{year}{2013}), \bibinfo{pages}{1108--1127}.
\newblock


\bibitem[\protect\citeauthoryear{Turner, Doxa, O'sullivan, and Penn}{Turner
  et~al\mbox{.}}{2001}]%
        {turner2001isovists}
\bibfield{author}{\bibinfo{person}{Alasdair Turner}, \bibinfo{person}{Maria
  Doxa}, \bibinfo{person}{David O'sullivan}, {and} \bibinfo{person}{Alan
  Penn}.} \bibinfo{year}{2001}\natexlab{}.
\newblock \showarticletitle{From isovists to visibility graphs: a methodology
  for the analysis of architectural space}.
\newblock \bibinfo{journal}{\emph{Environment and Planning B: Planning and
  design}} \bibinfo{volume}{28}, \bibinfo{number}{1} (\bibinfo{year}{2001}),
  \bibinfo{pages}{103--121}.
\newblock


\bibitem[\protect\citeauthoryear{Varoudis and Penn}{Varoudis and Penn}{2015}]%
        {varoudis2015visibility}
\bibfield{author}{\bibinfo{person}{Tasos Varoudis} {and} \bibinfo{person}{Alan
  Penn}.} \bibinfo{year}{2015}\natexlab{}.
\newblock \showarticletitle{Visibility, accessibility and beyond: next
  generation visibility graph analysis}. In \bibinfo{booktitle}{\emph{SSS
  2015-10th International Space Syntax Symposium}}.
\newblock


\bibitem[\protect\citeauthoryear{Varoudis and Psarra}{Varoudis and
  Psarra}{2014}]%
        {varoudis2014beyond}
\bibfield{author}{\bibinfo{person}{Tasos Varoudis} {and}
  \bibinfo{person}{Sophia Psarra}.} \bibinfo{year}{2014}\natexlab{}.
\newblock \showarticletitle{Beyond two dimensions: architecture through three
  dimensional visibility graph analysis}.
\newblock \bibinfo{journal}{\emph{The Journal of Space Syntax}}
  \bibinfo{volume}{5}, \bibinfo{number}{1} (\bibinfo{year}{2014}),
  \bibinfo{pages}{91--108}.
\newblock


\bibitem[\protect\citeauthoryear{Wald, Woop, Benthin, Johnson, and Ernst}{Wald
  et~al\mbox{.}}{2014}]%
        {Wald:2014:EKF:2601097.2601199}
\bibfield{author}{\bibinfo{person}{Ingo Wald}, \bibinfo{person}{Sven Woop},
  \bibinfo{person}{Carsten Benthin}, \bibinfo{person}{Gregory~S. Johnson},
  {and} \bibinfo{person}{Manfred Ernst}.} \bibinfo{year}{2014}\natexlab{}.
\newblock \showarticletitle{Embree: A Kernel Framework for Efficient CPU Ray
  Tracing}.
\newblock \bibinfo{journal}{\emph{ACM Trans. Graph.}} \bibinfo{volume}{33},
  \bibinfo{number}{4}, Article \bibinfo{articleno}{143} (\bibinfo{date}{July}
  \bibinfo{year}{2014}), \bibinfo{numpages}{8}~pages.
\newblock
\showISSN{0730-0301}
\urldef\tempurl%
\url{https://doi.org/10.1145/2601097.2601199}
\showDOI{\tempurl}


\end{thebibliography}

\appendix

\end{document}